\documentclass{article}

\usepackage{PRIMEarxiv}

\usepackage[utf8]{inputenc} 
\usepackage[T1]{fontenc}    

\usepackage{graphicx}
\usepackage{booktabs}
\usepackage{amsfonts}
\usepackage{multirow}
\usepackage{caption}
\usepackage{subcaption}
\usepackage{geometry}
\usepackage{mwe}
\usepackage{floatrow}
\usepackage{booktabs}       
\usepackage{amsfonts}       
\usepackage{nicefrac}       
\usepackage{microtype}      
\usepackage{fancyhdr}       
\usepackage{amsmath}
\usepackage{orcidlink}

\newcommand{\Lagr}{\mathcal{L}}

\title{Semantically Guided Representation Learning For Action Anticipation} 

\author{
  Anxhelo Diko\\
  Computer Science Department \\
  Sapienza University of Rome \\
  \texttt{diko@di.uniroma1.it}\\
  \And
  Danilo Avola\thanks{Equal second author contribution.}\\
  Computer Science Department \\
  Sapienza University of Rome \\
  \texttt{avola@di.uniroma1.it}\\
  \And
  Bardh Prenkaj\footnotemark[1]\\
  Chair of Responsible Data Science\\
  Technical University of Munich\\
  \texttt{bardh.prenkaj@tum.de}\\
  \And
  Federico Fontana\\
  Computer Science Department \\
  Sapienza University of Rome \\
  \texttt{fontana.f@di.uniroma1.it}\\
  \And
  Luigi Cinque\\
  Computer Science Department\\
  Sapienza University of Rome\\
  \texttt{cinque@di.uniroma1.it}
}

\begin{document}




\maketitle

\begin{abstract}
Action anticipation is the task of forecasting future activity from a partially observed sequence of events. However, this task is exposed to intrinsic future uncertainty and the difficulty of reasoning upon interconnected actions. Unlike previous works that focus on extrapolating better visual and temporal information, we concentrate on learning action representations that are aware of their semantic interconnectivity based on prototypical action patterns and contextual co-occurrences. To this end, we propose the novel Semantically Guided Representation Learning (S-GEAR) framework. S-GEAR learns visual action prototypes and leverages language models to structure their relationship, inducing semanticity. To gather insights on S-GEAR's effectiveness, we test it on four action anticipation benchmarks, obtaining improved results compared to previous works:  +3.5, +2.7, and +3.5 absolute points on Top-1 Accuracy on Epic-Kitchen 55, EGTEA Gaze+ and 50 Salads, respectively, and +0.8 on Top-5 Recall on Epic-Kitchens 100. We further observe that S-GEAR effectively transfers the geometric associations between actions from language to visual prototypes. Finally, S-GEAR opens new research frontiers in anticipation tasks by demonstrating the intricate impact of action semantic interconnectivity. \textcolor{blue}{\url{https://github.com/ADiko1997/S-GEAR}}

\end{abstract}


\section{Introduction}\label{ref:introduction}
Anticipating future actions is a key attribute of human intelligence for navigating the world. This remarkable skill translates directly to advanced computer vision applications such as self-driving cars \cite{funari@2021rolling,marchetti@2023mult} or wearable assistants \cite{zhong@2023anticipative,xu@2022dcr}, enabling safer navigation and better user experience \cite{qi@2023srl}. 
%
\begin{figure}[!t]
\centering
\includegraphics[width=\linewidth,viewport=0 0 1850 960]{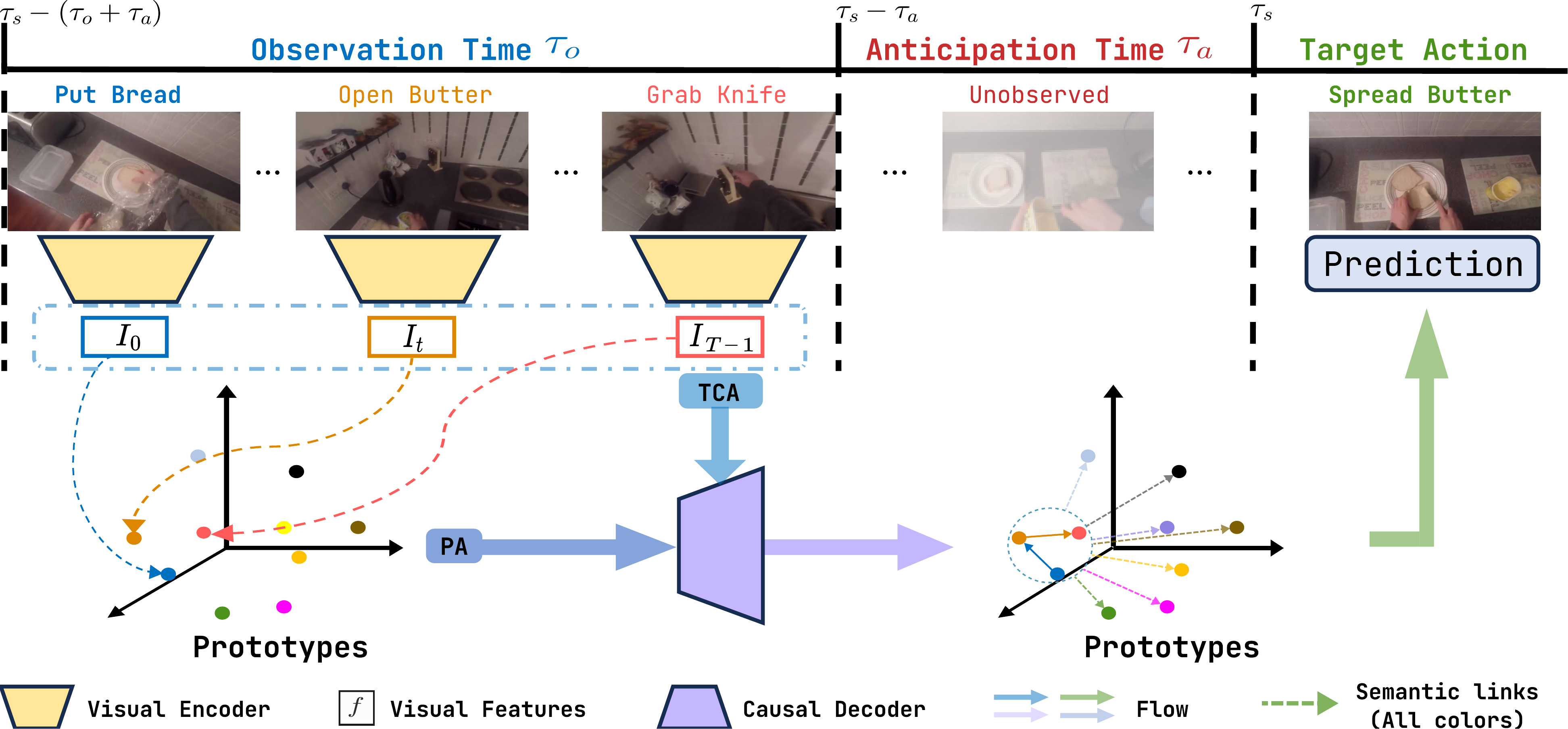}
 \caption{We propose learning action prototypes that encode typical action representations and meaningful semantic interconnections. The model leverages these prototypes to enhance the network encodings of observed actions and to forecast upcoming ones.}
 \label{fig:teaser}
\end{figure}

Recent developments in deep learning techniques have boosted the research on video understanding, reaching remarkable milestones on tasks like action recognition \cite{li2022mvitv2,wu@2022memvit,arnab2021vivit,girdhar2017attentional,liu2022swinV2,miech2017learnable}. Models related to action recognition can extrapolate essential spatiotemporal information from videos of isolated actions and correctly classify them. However, real-world applications operate in dynamic environments where actions are interconnected. For instance, imagine a self-driving car observing pedestrians. Predicting their intent to cross the street requires analyzing how observed dynamics relate to likely future events. This temporal misalignment between observation and future target introduces a challenge for recognition models to capture actionable insights, proving them insufficient and shifting the attention towards action anticipation \cite{funari@2021rolling,zhong@2023anticipative,girdhar@2021anticipative,vondrick2016anticipating,qi@2023srl,xu@2022dcr}.
This emerging research area focuses on enabling vision systems to predict future activity by observing ongoing events. A temporal gap between the observed events and the target future further amplifies its difficulty.

In trying to deal with the implications of action anticipation, previous methods extended recognition models with sequence units like LSTMs \cite{funari@2021rolling,qi@2023srl,xu@2022dcr,abu@2019uncertainty} and causal transformers \cite{girdhar@2021anticipative,xu@2022dcr,zhong@2023anticipative}.
The success of these approaches relies on the network's ability to effectively represent significant visual information from videos and the ability of the sequence units to propagate and preserve information in time. However, these methods have limitations. They cannot explicitly model the semantic connectivity between actions beyond the immediate video context, which is critical when dealing with co-occurring action sequences. According to cognitive sciences, semantic interconnectivity is fundamental for anticipating the future \cite{fredskin@2001propositional}. It helps structure our knowledge by associating actions with objects, intentions, and likely outcomes. This enables us to draw on past experiences to form reliable predictions even in unseen situations. Inspired by such observations, we raise the question: \textit{Is it possible to encode meaningful semanticity between action representations in a vision model?}

In pursuit of answering our question, we propose the \textit{\textbf{S}emantically \textbf{G}uided R\textbf{E}presentation Le\textbf{AR}ning} (\textbf{S-GEAR}) framework (see Fig. \ref{fig:teaser}). S-GEAR tackles action anticipation with a novel representation learning approach oriented by two fundamentals of actions semantic connectivity: \textbf{(1)} understanding the typical patterns of individual actions, and \textbf{(2)} modeling relationships between actions based on contextual co-occurrences \cite{ramanathan2015learning,wilson2017measuring,bullinaria2007extracting}. For \textbf{(1)}, S-GEAR learns a set of visual action prototypes. Each prototype encodes specific action patterns, capturing typical movements or gestures that define and distinguish action categories, reducing reliance on the specific appearance details of individual videos. Conversely, for \textbf{(2)}, building semantic relationships between actions solely from videos is challenging. First, it requires processing long action sequences to include enough context and defining co-occurrence relationships. Second, actions are usually not represented equally in videos \cite{damen2018scaling,dima2022rescaling}, hardening the modeling of under-represented action relationships. S-GEAR circumnavigates these issues by exploiting language models known to extract inter-concept semantic relationships  \cite{somin@2023relation,reimers2019sentence} -- i.e., language models effectively tackle \textbf{(2)}. Specifically, S-GEAR creates language prototypes based on action labels and transfers their inherent semantic connectivity to visual prototypes without aligning them directly. To achieve this, S-GEAR uses a new loss function that enables visual prototypes to maintain visual cues, such as object and movement patterns, while encoding semanticity by mimicking the geometric associations between actions from language.

S-GEAR uses an encoder-decoder transformer architecture to learn prototypes and encode semantic relationships between actions. The encoder consists of a standard \textit{Vision Transformer} (ViT) \cite{dosovitskiy2021ViT,ashish@2017attention} for visual context, while the decoder is a \textit{Causal Transformer} (CT) \cite{wang2019learning,ashish@2017attention} which models temporal causality. These structures are interconnected through two novel computational blocks, namely \textit{Temporal Context Aggregator} (TCA) and \textit{Prototype Attention} (PA) for, respectively, causality enhancement and semanticity promotion. Lastly, S-GEAR appends a classification head that produces future class probabilities based on the decoder's output and geometric association with the visual prototypes.

To assess S-GEAR's performance, we conduct extensive experiments on two egocentric video datasets, Epic-Kitchens \cite{damen2018scaling,dima2022rescaling} (both Epic-Kitchens 55 and Epic-Kitchens 100 versions) and EGTEA Gaze+ \cite{zhang@2019gaze}. Moreover, we evaluate S-GEAR on an exocentric dataset, namely 50 Salads \cite{stein2013combining}, to demonstrate its versatility in tackling long-term dense anticipation. We show that S-GEAR improves over the current state-of-the-art in most scenarios. We also conduct ablation studies highlighting the usefulness of the semantic connectivity between the actions that S-GEAR incorporates. 

This paper's contributions are fourfold. \textbf{(1)} We present S-GEAR, a novel prototype learning framework for action anticipation leveraging action interconnectivity. \textbf{(2)} We introduce a novel approach to map semanticity from language to vision without direct alignment between modalities. \textbf{(3)} We conduct extensive experiments on two egocentric datasets and an exocentric one to highlight S-GEAR's versatility in different action anticipation scenarios (i.e., egocentric vs. exocentric and short-term vs. long-term). \textbf{(4)} We showcase the benefits of S-GEAR w.r.t. its counterparts that do not rely on semantic relationships.

\section{Related Work}\label{ref:related_work}



\noindent\textbf{Action anticipation} predicts future actions before they occur in video clips and is well explored both in third-person (exocentric) videos
\cite{koppula2015anticipating,vondrick2016anticipating,singh2016krishnacam,soran2015generating,bokhari2017long}, and first-person (egocentric) videos \cite{funari@2021rolling,furnari2018leveraging,qi@2023srl,zhang@2019gaze,dessalene@2023omg,xu@2022dcr,roy2022action,girdhar@2021anticipative,girase@2023latency,liu2022hybrid,roy2024interaction}, due to its applicability on autonomous agents and wearable assistants \cite{defeng@2023seem,marchetti@2023mult,girdhar@2021anticipative}. 
Funari et al. \cite{funari@2021rolling} introduce RU-LSTM, a model with two LSTMs and a modality attention component. Osman et al. \cite{osman@2021slowfast} integrate RU-LSTM into SlowFast. Qi et al. \cite{qi@2023srl} enhance LSTMs with Self-Regulated Learning (SRL). Dessalene et al. \cite{dessalene@2023omg} use hand-object contact representations for action anticipation. Xu et al. \cite{xu@2022dcr} employ curriculum learning. Roy et al. \cite{roy2022action} predict final goals for near-future anticipation. Liu et al. \cite{liu2022hybrid} store long-term action prototypes for richer short-term representations. Girdhar et al. \cite{girdhar@2021anticipative} propose AVT, combining ViT and causal transformer, paving the way for \cite{zhong@2023anticipative,girase@2023latency,wu@2022memvit}.
Unlike previous works, S-GEAR considers the semantic relationship between action representations \cite{fredskin@2001propositional} by using vision and language prototypes to guide the model's training process semantically.

\noindent\textbf{Vision-Language alignment} relies on effectively aligning concepts between vision and language in a unified representation space. Typically achieved through contrastive training of modality encoders \cite{zhai2022lit,jia2021scaling,radford2021learning}, these methods use vision-language pairs for encouraging proximity between corresponding visual and text embeddings. Zhai et al. \cite{zhai2022lit} utilize contrastive learning to align text encoder representations with a frozen pre-trained vision model. Radford et al. \cite{radford2021learning} introduce CLIP, training separate encoders for text and images and aligning representations through contrastive loss. Ma et al. \cite{ma2022x} extend CLIP to videos, employing multi-grained contrastive learning. Advancements include cross-modal fusion architectures using a cross-modality encoder for text and visual inputs \cite{akbari2021vatt,huang2023clover}.
Unlike previous works, S-GEAR only translates the geometric association between action prototypes from language to vision without shifting spaces. 

\noindent\textbf{Prototype Learning} involves creating characteristic ``prototypes'' of labeled data samples. Initially dominant in few-shot learning for novel class prediction \cite{sung2018learning,snell2017prototypical}, this strategy now successfully encodes spatial and temporal patterns in domains such as video semantic segmentation \cite{lin2023@prototypical} and action recognition \cite{martinez2019action}.
\section{Method}\label{ref:method}

%
We propose S-GEAR for action anticipation. S-GEAR discerns essential spatiotemporal signals and understands the semantic relationships between actions. It contains a neural network architecture tailored for understanding spatiotemporal video sequences and a learning policy that guides the network semantically to map out the interconnections between actions.

\noindent\textbf{Task Formulation.} Action anticipation involves predicting an action category for an event starting at time $\tau_s$, observing a video segment $V_o$ within the interval $[\tau_s - (\tau_o + \tau_a); \tau_s - \tau_a]$ \cite{dima2022rescaling}. Here, $\tau_o$ and $\tau_a$ denote the observation and anticipation periods set specifically to the dataset.

\begin{figure}[!t]
\centering  
\includegraphics[width=\linewidth,viewport=0 0 2800 960]{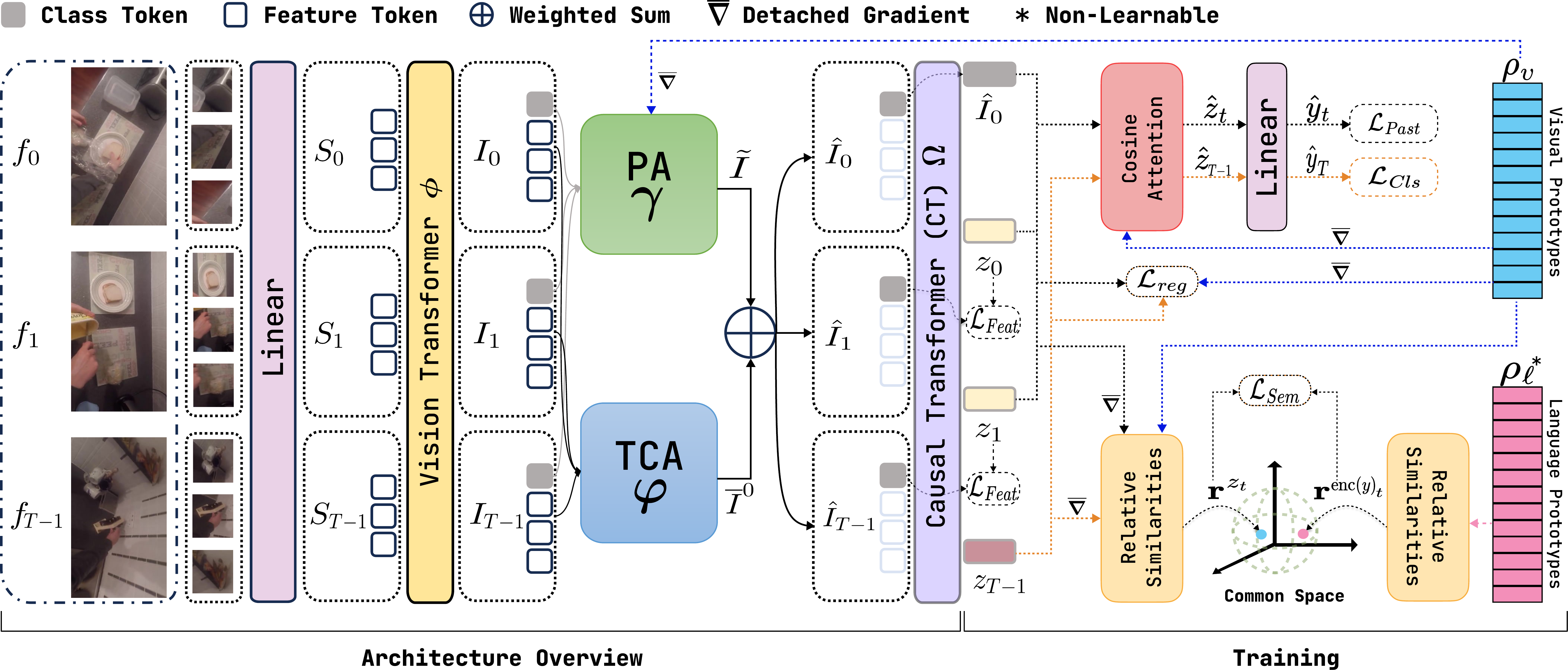}
 \caption{
 S-GEAR processes frame sequence patches and creates input token sequences $S_t$. ViT $\phi$ encodes $S_t$ into intermediate features $I_t$. PA $\gamma$ and TCA $\varphi$ process $I_t$, merging outputs into semantically enhanced causal features $\hat{I}_t$. Class tokens pass through the CT decoder $\Omega$, predicting future features $z_{t}$. The features $z_t$  and the proposed prototypes are trained for action anticipation ($\Lagr_{Cls}$) and semantic relation encodings ($\Lagr_{Sem}$). The network is also regularized for accurate future representations ($\Lagr_{Feat}$) and correct past action classification ($\Lagr_{Past}$). Finally, a distance loss ($\Lagr_{reg}$) is applied to $z_t$.
 }
 \label{fig:fig_3}
\end{figure}

\subsection{Proposed Architecture}\label{ss:architecture_overview}

S-GEAR processes a sequence of video frames and produces a set of features that can accurately describe the subsequent action. To achieve this, as shown in Fig. \ref{fig:fig_3}, S-GEAR employs an architecture composed of (1) a visual encoder for extracting feature vectors from the input frames; (2) the Temporal Context Aggregator (TCA) module designed to incorporate detailed temporal context from past to current observations; (3) the Prototype Attention (PA) block, which combines visual features with learned prototypes and (4) the Causal Transformer (CT) decoder responsible for predicting future representations. 

\noindent\textbf{Visual Encoder.} Upon receiving a video segment $V_o=\{f_0,...,f_{T-1}\}$ of length $T$, S-GEAR relies on ViT \cite{dosovitskiy2021ViT} as the visual encoder $\phi$ to obtain spatial features from each frame. ViT splits each frame into $P$ non-overlapping patches of equal size, which are then flattened and transformed into a series of feature tokens $S_t \in \mathbb{R}^{P \times d}$ corresponding to frame $f_t \in V_o$. Here, $d$ represents the token dimensionality. Then, to preserve the spatial context, learnable positional encodings are added to $S_t$. Additionally, the so-called ``\textit{class token}'' $\text{CLS}_t$, which captures the global context of frame $f_t$, is prepended to $S_t$. The transformer blocks then act on $S_t$,  generating visual features $I_t=\phi(S_t)$ with the same dimension as $S_t$.

\noindent\textbf{Temporal Context Aggregator (TCA) and Prototype Attention (PA).} In this stage, $I_t$ passes through two specialized units to enhance temporal causality and semantic interconnections between actions. Inspired by the left-to-right causal transformer \cite{ashish@2017attention}, we craft TCA $\varphi$, the first unit, to effectively transfer comprehensive context from the past to the current frame representation. In TCA, unlike standard causal blocks that mainly rely on the global representations $I^0_t$, we consider all feature patches in each frame. Thus, given all global and local representations $I$ of the frames, we obtain causal intermediate features $\overline{I} \in \mathbb{R}^{T\times (P+1) \times d}$ where each $\overline{I}_t \in \overline{I}$ is enhanced with detailed contextual information from past frames. Contrarily, the second unit PA, denoted as $\gamma$, operates parallel to the TCA on $I_t^0$ and the visual prototypes. Specifically, PA aggregates information from selected visual prototypes upon feature similarity to $I_t^0$, promoting semantic relation encoding between actions as inferred from the different prototypes. We rely on the attention mechanism using $I^0_t$ as queries and the visual prototypes as keys and values to produce semantically enhanced feature sets $\Tilde{I} \in \mathbb{R}^{T\times d}$. We then combine $\Tilde{I}$ and $\overline{I}^0$ as a weighted sum $\hat{I} = \lambda \overline{I}^0 + (1-\lambda)\Tilde{I}$ ($\lambda$ is learnable). We point the reader to Appendix A.1-2 for details on TCA and PA.
 
\noindent\textbf{Temporal Decoder.} We rely on an autoregressive Causal Transformer (CT) decoder $\Omega$, as presented in \cite{xu@2022dcr,girdhar@2021anticipative,zhong@2023anticipative} to analyze $\hat{I}$ from $t=0$ to $t=T-1$ and generate a set of features that describes the likely future. Similar to the visual encoder, we add learnable positional encodings to $\hat{I}$ to preserve the temporal context. Afterward, we feed the embedded features with positional encodings to the decoder blocks, built upon the masked multi-head self-attention \cite{girdhar@2021anticipative}. Thus, $\Omega$ generates a new sequence $\zeta = \Omega(\hat{I})$ s.t. $\forall t$, $z_t \in \zeta$ represents the future features of $\hat{I}_t$ after observing all the past ones including itself. For $t=T-1$, $z_t$ represents the future action happening $\tau_a$ seconds after the observed sequences.

\subsection{Semantic Guiding Policy}\label{ss:semantic_guiding_policy}

We exploit vision/language prototypes and a common communication space between them to facilitate a semantic-based guiding policy for action anticipation.

\noindent\textbf{Prototypes.} We aim to translate semantic relationships from language-based action concepts to the visual domain. Thus, we define two sets of prototypes. The first, defined as the language prototypes $\rho_\ell \in \mathbb{R}^{K\times d}$ (where $K$ is the number of action classes), is extracted by encoding action labels composed of verb and noun combination using the ``Sentence Transformer'' proposed in \cite{reimers2019sentence}. These prototypes serve as the reference space for learning actions ``semantic connectivity'' \cite{somin@2023relation}. The second, defined as the visual prototypes $\rho_\upsilon \in \mathbb{R}^{K \times d}$, ensures that S-GEAR remains in the visual domain and effectively preserves characteristic visual patterns. Such prototypes are learnable and initialized from typical action samples encoded from the proposed architecture trained for action recognition. We exploit $\rho_\upsilon$ to encode visual action representations and inherit the semanticity from $\rho_\ell$. Refer to Appendix A.3 for initialization details.

\noindent\textbf{Common Communications Space.} To translate action relationships from language to vision without shifting domains, we define a common space where vision and language representations co-exist and are compared via their relative associations w.r.t. the prototypes. In more detail, given an action \textit{visual encoding} $z_t \in \zeta$, we compute its relative representation by comparing it against all elements in $\rho_\upsilon$ using a similarity function: i.e., $\mathbf{r}^{z_t} = \{r_1^{z_t},\dots,r_{K}^{z_t}\}$ s.t. $r_k^{z_t} = cos(z_t,\rho_\upsilon[k])$ for each action class $k \in \{1,\dots, K\}$. Similarly, we compute the relative representation of a \textit{language encoding} $\text{enc}(y)_t$ -- i.e., the language encoding of the action label at time $t$ -- against the prototypes in $\rho_\ell$ as $\mathbf{r}^{\text{enc}(y)_t} = \{r_1^{\text{enc}(y)_t},\dots,r_K^{\text{enc}(y)_t}\}$ s.t. $\mathbf{r}_k^{\text{enc}(y)_t} = cos(\text{enc}(y)_t,\rho_\ell[k])$ for all action classes $k \in \{1,\dots, K\}$. Now, we ensure that each $k$-th entry in $\mathbf{r}^{z_t}$ and $\mathbf{r}_k^{\text{enc}(y)_t}$ represents the geometric association with the $k$-th class prototype in the language/vision domain. Hence, we can directly compare these two representations based on their relative position in their original vector spaces.

\subsection{Training}\label{ss:training}

To train the model, for each labeled action segment, we sample a clip preceding it and ending exactly $\tau_a$ seconds before the start of the action. We pass the clip through S-GEAR to obtain $z_t$ and then optimize it to learn semantically and visually meaningful prototypes for action anticipation. 

\noindent\textbf{Prototype Learning.}\label{Prototype_learning}
Learning prototypes aim to establish a visual latent space where predefined semantic connections describe actions by ``aligning'' the latent space topology defined by $\rho_\upsilon$ with $\rho_\ell$. To do so, we calculate the relative positions\footnote{$z_t$ is detached from the gradient when calculating $\mathbb{\textbf{r}}^{z_t}$ to avoid collapsing issues.} $\mathbb{\textbf{r}}^{z_t}$ and $\mathbf{r}^{\text{enc}(y)_t}$, which we use to define the semantic loss in Eq. \ref{eq:sem_loss}.
\begin{equation}\label{eq:sem_loss}
    \Lagr_{Sem} = \big|\mathbb{\textbf{r}}^{z_t} - \mathbf{r}^{\text{enc}(y)_t}\big|.
\end{equation}
During optimization, the prototypes in $\rho_\upsilon$ will be refined to represent relative relationships between actions akin to those inferred from the language space. Additionally, to guide S-GEAR push the action $z_t$ towards the prototype of the same class $k$ (i.e., $\rho_\upsilon[k]$) and avoid divergences, we add a lasso regularization\footnote{$\rho_\upsilon[k]$ is detached from the gradient to avoid collapsing issues during regularization.} to $\Lagr_{Sem}$ as in Eq. \ref{eq:vision_reg}.
\begin{equation}\label{eq:vision_reg}
\begin{gathered}
    \Lagr_{reg} = \big|\big|z_t - \rho_\upsilon[k]\big|\big|_2^2\\
    \Lagr_{Sem} = \Lagr_{Sem} + \Lagr_{reg}.
\end{gathered}
\end{equation}
Thus, while shaping the visual latent space geometry defined by $\rho_\upsilon$ (Eq. \ref{eq:sem_loss}), we enforce action representations to fall close to their visual prototype (Eq. \ref{eq:vision_reg}). 

\noindent\textbf{Anticipation Training.}
Besides prototype learning, we train S-GEAR for action anticipation by optimizing the cross-entropy loss between the predicted class label $\hat{y}_T$ and the ground truth $y_T$. $\hat{y}_{T}$ is obtained from the encoded action representation and its relative position w.r.t. the visual prototypes. More specifically, for the action representation $z_{T-1}$, we calculate $\mathbf{r}^{z_{T-1}}$. Since $\mathbf{r}^{z_{T-1}}$ contains values in $[-1,+1]$, we transform them into probabilistic weights using softmax. Now, we aggregate all the prototype vectors into a single representation, $\overline{z}_{T-1} \in \mathbb{R}^d$, according to the obtained weights (see Eq. \ref{eq:softmaxed_prototypes}).
\begin{equation}\label{eq:softmaxed_prototypes}
    \overline{z}_{T-1} = \text{softmax}(\mathbb{\textbf{r}}^{z_{T-1}})\cdot \rho_\upsilon.
\end{equation}
To jointly learn the action representation and its exact collocation in the visual space w.r.t. the prototypes, we perform a weighted sum as in Eq. \ref{eq:weighted_sum_action_repr}:
\begin{equation}\label{eq:weighted_sum_action_repr}
    \hat{z}_{T-1}=\sigma(\alpha) z_{T-1} + (1-\sigma(\alpha)) \overline{z}_{T-1},
\end{equation}
where $\sigma$ is a sigmoid function, and $\alpha$ is a learnable scalar. Such operations are represented as Cosine Attention in Fig. \ref{fig:fig_3}. Lastly, we feed $\hat{z}_{T-1}$ through a linear layer and softmax its output to obtain $\hat{y}_{T}$. We calculate the cross-entropy loss (Eq. \ref{eq:ce}) between the ground truth and the predicted action class.
\begin{equation}\label{eq:ce}
    \Lagr_{Cls} = -\sum_{i}^{K}y_{T}^{i}\log(\hat{y}_{T}^{i}).
\end{equation}
\noindent Additionally, inspired by \cite{girdhar@2021anticipative,vondrick2016anticipating}, we leverage the causality of the decoder $\Omega$. Here, we use any true class label for the past frames and minimize the cross-entropy on past label predictions (Eq. \ref{eq:ce_past}). Notice that the predicted label $\hat{y}_t$ is produced following the same reasoning described above for $\hat{y}_{T}$ (see Eq. \ref{eq:softmaxed_prototypes}, \ref{eq:weighted_sum_action_repr}).
\begin{equation}\label{eq:ce_past}
    \Lagr_{Past} = -\sum_{t=0}^{T-2}\sum_{i}^{K}y_{t+1}^{i}\log(\hat{y}_{t+1}^{i}).
\end{equation}
To produce faithful future features, we minimize the distance between the predicted future frame features and the actual ones:
\begin{equation}\label{eq:feats}
    \Lagr_{Feat} = \sum_{t=0}^{T-2}\big|\big|\hat{I}_{t+1} - z_t\big|\big|. 
\end{equation}
The overall loss function used to train S-GEAR is a weighted sum of all the individual losses: $\mathcal{L}_{tot} = \lambda_1\mathcal{L}_{Sem}+\lambda_2\mathcal{L}_{Cls}+\lambda_3\mathcal{L}_{Past}+\lambda_4\mathcal{L}_{Feat}$.
\section{Experiments}\label{ref:experiments}

\subsection{Datasets and Metrics}\label{ss:dataset_metric}
 \textbf{The EPIC-Kitchens 55} (EK55) dataset \cite{damen2018scaling} is a medium-scale first-person cooking dataset comprising 432 videos from 32 different individuals and approximately 40,000 segments. It encompasses 92 verbs and 272 object classes, resulting in 2,747 action classes. Additionally, we use the train and validation splits provided in \cite{funari@2021rolling}. Our model's performance on EK55 is evaluated using Top-1/5 Accuracy at $\tau_a = 1$s, following prior works \cite{funari@2021rolling,qi@2023srl,xu@2022dcr,girdhar@2021anticipative}.
 
 \noindent\textbf{The EPIC-Kitchens 100} (EK100) dataset \cite{dima2022rescaling} is a substantial extension of EK55, encompassing 700 videos from 37 individuals in 45 diverse kitchens. It comprises $\sim$90,000 activity segments spanning 495 training, 138 validation, and 67 test videos. EK100 offers a richer representation of cooking activities through its broader range of verbs (97), objects/nouns (300), and action classes (4,053). To assess model performance on EK100, we employ the class aware mean Top-5 Recall \cite{dima2022rescaling,girdhar@2021anticipative,xu@2022dcr} metric at $\tau_a = 1$s. 

\noindent\textbf{The EGTEA Gaze+} dataset (EG) \cite{zhang@2019gaze} includes 28 hours of first-person cooking videos from 32 subjects across 86 sessions, covering 7 tasks. The dataset contains 10,325 activity instances, categorized into 19 verbs, 51 objects, and 106 activity classes. To evaluate our model, we employed Top-1 Accuracy on split 1 for $\tau_a=0.5$s \cite{abu@2018will,zhong@2023anticipative,girdhar@2021anticipative} and Top-5 Accuracy averaged across all three splits to evaluate overall performance for $\tau_a=1$s \cite{funari@2021rolling,qi@2023srl,liu2022hybrid}.  

\noindent\textbf{The 50 Salads} dataset (50S) \cite{stein2013combining} comprises 50 exocentric videos featuring salad preparation activities performed by 25 different actors and categorized into 17 activity classes. We assess our model using mean Top-1 Accuracy across the 5 official splits following previous works \cite{qi@2023srl,funari@2021rolling}. Unlike other benchmarks, the 50S offers a dense action anticipation challenge with variable observation and anticipation times. Specifically, for a given video segment in input, $\tau_a$ goes from $10\%$ to $50\%$ of the video's duration while $\tau_o$ is set to $20\%$ or $30\%$.

\subsection{Implementation Settings}

\noindent\textbf{Visual Encoder.} S-GEAR employs the ViT Base (ViT-B) architecture as its visual encoder with a patch size of 16$\times$16. It comprises 12 transformer blocks, feature dimension 768, and operates with 12 attention heads. We set each frame size for input dimensions to 384$\times$384 for the EK55/100 datasets and 224$\times$224 for the EG and 50S datasets. Besides the default encoder, following prior works \cite{xu@2022dcr,qi@2023srl,girdhar@2021anticipative}, we show that S-GEAR can also be used with other backbones like TSN and irCSN using pre-extracted features as in \cite{funari@2021rolling} and \cite{girdhar@2021anticipative}, respectively.

\noindent\textbf{Intermediate Stage.} Our intermediate processing stage, crucial for linking the visual encoder's output to the causal transformer decoder, consists of 2 TCA blocks and 1 PA block. Note that when replacing ViT with other backbones, we omit TCA blocks. This is because, without ViT's detailed local patches, the architecture essentially becomes a standard causal transformer.

\noindent\textbf{Causal Transformer Decoder.} 
For EK55/100 datasets, we employ a 6-layer causal transformer decoder with 4 heads and a dimensionality of 2048 to process the observed context and predict future events. For the EG dataset, we reduce the number of layers to 2. Meanwhile, for the 50S dataset, an 8-layer decoder with eight heads and the same dimensionality is used.

\noindent\textbf{Observation.} 
For EK100, we set the observation time, $\tau_o$, to 15s, processing video segments at 1fps. For EK55 and EG, we maintain the same processing rate but reduce $\tau_o$ to 10s. In contrast, for the 50S, we align with \cite{qi@2023srl,ke@2019time,abu@2019uncertainty} and adopt observation rates of 20\% and 30\% for each input sequence, with 0.25fps.

\noindent\textbf{Training Settings.}
We employ different training strategies for each dataset. For EK100, EK55, and EG, we use an SGD optimizer with a momentum of 0.9 and weight decay of 1e-5, processing mini-batches of 3. The learning rates are 1e-4 for EK55/100 and 4.75e-4 for EG, all with cosine scheduling and warmup of 10, 20, and 5 epochs, respectively. The total training durations are 50 epochs for EK100, 35 for EK55, and 10 for EG. In contrast, for 50S, we opt for AdamW optimizer with parameters $\beta_1$, $\beta_2$ set to 0.9, 0.999, a weight decay of 1e-4, and a learning rate 5e-6. This setup also includes cosine scheduling and 20 warmup epochs, with the model training for 100 epochs on mini-batches of 2.
Finally, we run our experiments on an RTX4090 and 2$\times$V100 GPUs.

\subsection{Baselines}
We compare against 
RU-LSTM \cite{funari@2021rolling}, SRL \cite{qi@2023srl}, AVT \cite{girdhar@2021anticipative}, DCR \cite{xu@2022dcr}, MeMViT \cite{wu@2022memvit}, RAFTformer \cite{girase@2023latency}, HRO \cite{liu2022hybrid}, AFFT \cite{zhong@2023anticipative}, TempAgg. \cite{sener2020temporal}, Imagination \cite{wu2020learning} and more to ensure a fair comparison.
Bold and underlined values in the tables illustrate the best and second-best results, respectively.
\begin{table}[!t]
    \caption{Experiments on Epic-Kitchens 55/100 for $\tau_a$=1s.}
    \begin{subtable}[c]{.47\linewidth}\vspace{-2pt}
    \centering
    \caption{Unimodal results on EK55 validation set.}
    \label{tab:st_ek55_uni}
    \resizebox{\linewidth}{!}{%
        \begin{tabular}[c]{c | l c c c c}
             \toprule
             \parbox[t]{3mm}{\multirow{14}{*}{\rotatebox[origin=c]{90}{RGB}}} & Model & Encoder & Initialization & Top-1 Acc. & Top-5 Acc. \\ \midrule
             & RU-LSTM \cite{funari@2021rolling} & TSN &IN1K& 13.1 & 30.8\\
             & SRL \cite{qi@2023srl} & TSN &IN1K& / & 31.7 \\
             & AVT \cite{girdhar@2021anticipative} & TSN &IN1K& 13.1 & 28.1 \\
             & AVT \cite{girdhar@2021anticipative} & ViT-B &IN21K& 12.5 & 30.1 \\
             & AVT \cite{girdhar@2021anticipative} & irCSN &IG65M& 14.4 & 31.7 \\
             & DCR \cite{xu@2022dcr} & TSN &IN1K& 13.6 & 30.8 \\
             & DCR \cite{xu@2022dcr} & irCSN &IG65M& 14.4 & \underline{34.0} \\
             & DCR \cite{xu@2022dcr} & TSM &K400& \underline{16.1} & 33.1 \\
             & \textbf{S-GEAR (ours)} & TSN &IN1K& 15.6 & 32.8 \\  
             & \textbf{S-GEAR (ours)} & irCSN &IG65M& \textbf{16.2} & 33.1\\ 
             & \textbf{S-GEAR (ours)} & ViT-B &IN21K& 15.8 & \textbf{34.5} \\ 
             \midrule
             \parbox[t]{3mm}{\multirow{3}{*}{\rotatebox[origin=c]{90}{Obj}}} 
             & RULSTM & FRCNN &IN1K& 10.0 & 29.8\\ 
             & DCR & FRCNN &IN1K& \underline{11.5} & \textbf{30.5}\\ 
             & \textbf{S-GEAR (ours)} & FRCNN &IN1K& \textbf{12.45} & \underline{30.4}\\ 
             \midrule
             \parbox[t]{3mm}{\multirow{3}{*}{\rotatebox[origin=c]{90}{Flow}}}
             & RULSTM & TSN &IN1K& 8.7 & 21.4\\ 
             & DCR & TSN &IN1K& \underline{8.9} & \underline{22.7} \\ 
             & \textbf{S-GEAR (ours)} & TSN &IN1K& \textbf{10.8} & \textbf{25.8} \\
            \bottomrule
        \end{tabular}%
        }
    \end{subtable}
    %
    \begin{subtable}[c]{.5\linewidth}
    \centering
    \caption{Unimodal results on EK100 validation set.}
    \label{tab:st_ek100_uni}
    \resizebox{\linewidth}{!}{%
    \begin{tabular}{c|l c cccc}
         \toprule
         \parbox[t]{3mm}{\multirow{15}{*}{\rotatebox[origin=c]{90}{RGB}}} & Model & Encoder & Initialization & Verb & Noun & Action \\ \midrule
         & RU-LSTM \cite{funari@2021rolling} & TSN &IN1K& / & / & 13.3 \\
         & AVT \cite{girdhar@2021anticipative} & ViT &IN21K& 30.2 & 31.7 & 14.9 \\
         & DCR \cite{xu@2022dcr} & TSM &K400& 32.6 & 32.7 &  16.1 \\
         & MeMViT \cite{wu@2022memvit} & MViTv2-16 &K400& 32.8 & 33.2 & 15.1 \\    
         & MeMViT \cite{wu@2022memvit} & MViTv2-24 &K700& 32.2 & 37.0 & 17.7 \\    
         & RAFTformer \cite{girase@2023latency} & MViTv2-16 &K400& 33.3 & 35.5 & 17.6 \\
         & RAFTformer \cite{girase@2023latency} & MViTv2-24 &K700& \underline{33.7} & 37.1 & 18.0 \\
         & RAFTformer-2B \cite{girase@2023latency} & MViTv2-16\&24 &K400\&700& \textbf{33.8} & \underline{37.9} & \underline{19.1}\\
         & \textbf{S-GEAR (ours)} & ViT-B &IN21K& 31.1 & 37.3 & 18.3 \\
         & \textbf{S-GEAR (ours)} & TSN &IN1K& 25.8 & 29.8 & 14.9 \\
         & \textbf{S-GEAR (ours)} & irCSN &IG65M& 26.8 & 28.8 & 13.3 \\
         & \textbf{S-GEAR-2B (ours)} & ViT-B$\times$2 &IN21K& 32.7 & \textbf{37.9} & \textbf{19.6} \\
         \midrule
         \parbox[t]{3mm}{\multirow{3}{*}{\rotatebox[origin=c]{90}{Obj}}} 
         & AVT & FRCNN &IN1K& 18.0 & \underline{24.3} & 8.7 \\ 
         & DCR & FRCNN &IN1K& \textbf{22.2} & 24.2 & \underline{9.7}  \\ 
         & \textbf{S-GEAR (ours)} & FRCNN &IN1K& \underline{20.8} & \textbf{28.6} & \textbf{11.4} \\ 
         \midrule
         \parbox[t]{3mm}{\multirow{3}{*}{\rotatebox[origin=c]{90}{Flow}}}
         & AVT & TSN &IN1K& 20.9 & 16.9 & 6.6\\ 
         & DCR & TSN &IN1K& \textbf{25.9} & \underline{17.6} & \textbf{8.4} \\ 
         & \textbf{S-GEAR (ours)} & TSN &IN1K& \underline{21.5} & \textbf{18.2} & \underline{7.9} \\
         \bottomrule
    \end{tabular}%
    }
    
    \end{subtable}
    \begin{subtable}[c]{.45\linewidth} 
    \centering
    \caption{Multimodal results on EK55 validation set.}
        \label{tab:st_ek55_multi}
    \resizebox{\linewidth}{!}{%
        \begin{tabular}[l]{l cc c c}
             \toprule
             Model &  Modalities & & Top-1 Acc. & Top-5 Acc. \\ \midrule
             
             RU-LSTM  & RGB+Obj+Flow && 15.3 & 35.3\\
             TempAgg.  & RGB+Obj+Flow && 15.1 & 35.6 \\
             Imagination & RGB+Obj+Flow && 15.2 & 35.4 \\ 
             SRL & RGB+Obj+Flow &&  / & 35.5 \\
             Ego-OMG & RGB+HOI+NAO && {\color{gray}19.2} & / \\
             AVT+ \cite{girdhar@2021anticipative}& RGB+Obj && 16.6 & 37.6 \\
             HRO & RGB+Obj+Flow && / & 37.4 \\
             DCR & RGB+Obj+Flow && \underline{19.2} & \underline{41.2} \\
             \textbf{S-GEAR (ours)}& RGB+Obj+Flow && \textbf{22.7} & \textbf{43.2} \\ \bottomrule
        \end{tabular}%
        }
    \end{subtable}
    \hfill
    \begin{subtable}[c]{.5\linewidth}
    \centering
    \caption{Multimodal results on EK100 validation and test sets. HOI refers to Hand-Object-Interaction.}
    \label{tab:st_ek100_multi}
    \resizebox{\linewidth}{!}{%
    \begin{tabular}[r]{l cc ccc c ccc}
         \toprule
         \multirow{2}{*}{Model}& \multirow{2}{*}{Modalities} & &\multicolumn{3}{c}{Validation} && \multicolumn{3}{c}{Test} \\  \cmidrule{4-6} \cmidrule{8-10}
         & && Verb & Noun & Action && Verb & Noun & Action \\ \midrule
         RU-LSTM & RGB+Obj+Flow && 27.8 & 30.8 & 14.0 && 25.3 & 26.7 & 11.2 \\
         TempAgg. & RGB+Obj+Flow+HOI+Audio && 23.2 & 31.4 & 14.7 && 21.8 & 30.8 & 12.6 \\
         AVT+ & RGB+Obj && 28.2 & 32.0 & 15.9 && 25.6 & 28.8 & 12.6 \\
         AFFT & RGB+Obj+Flow+HOI+Audio && 22.8 & 34.6 & 18.5 && 20.7 & 31.8 & 14.9\\
         \textbf{S-GEAR (ours)}& RGB+Obj && 29.5 & \underline{37.8} & 18.9 && \underline{25.9} & 32.0 & 14.7 \\ 
         \textbf{S-GEAR-2B (ours)}& RGB+Obj && \textbf{30.5} & \textbf{38.4} & \underline{19.6} && 25.5 & \underline{31.7} & \underline{15.3} \\ 
         \textbf{S-GEAR-4B (ours)}& RGB+Obj && \underline{30.2} & 37.0 & \textbf{19.9} && \textbf{26.6} & \textbf{32.6} & \textbf{15.5}\\ 
         \bottomrule
    \end{tabular}%
    }
    \end{subtable}
    \label{tab:ek_experiments}
\end{table}

\subsection{Unimodal Comparison}
Table \ref{tab:ek_experiments} (a), (b) provide unimodal results on EK55 and EK100 datasets, ensuring a fair comparison of S-GEAR against baselines. In EK55 (Table \ref{tab:ek_experiments} (a)), in RGB, S-GEAR demonstrates a point improvement of 1.1 on Top-5 Acc. (vs. the second-best SRL) and 2.0 on Top-1 Acc. (vs. the second-best DCR) for the TSN features. Regarding the irCSN features, S-GEAR surpasses DCR by 1.8 points in Top-1 Acc. while trailing it on Top-5 Acc. by 0.9. Using the ViT-B backbone, S-GEAR surpasses AVT  by 3.3 (Top-1) and 6.4 (Top-5) 
although S-GEAR uses bigger frame sizes. 
For the object modality, we use Faster R-CNN features (as in \cite{funari@2021rolling,qi@2023srl,xu@2022dcr,girdhar@2021anticipative}) for a fair comparison, obtaining 0.9 Top-1 Acc. improvement, yet failing behind on Top-5 by 0.1. Finally, S-GEAR yields 1.9 (Top-1) and 3.1 (Top-5) point gains for the flow modality over prior works. 

Table \ref{tab:ek_experiments} (b) details our results on the more complex EK100 benchmark. Here, S-GEAR competes with MeMViT \cite{wu@2022memvit} and RAFTformer \cite{girase@2023latency} (with the MViTv2-16 backbone) on the RGB modality. S-GEAR demonstrates improvements in Top-5 Recall for actions (3.2 over MeMViT, 0.7 over RAFTformer) and nouns (4.1 over MeMViT, 1.8 over RAFTformer). While trailing slightly on verbs, unlike its competitors (Kinetics-400), S-GEAR performs well without spatiotemporal initialization. Surprisingly, even when MeMViT and RAFTformer employ larger backbones with Kinetics-700 initialization, S-GEAR exceeds them on actions and nouns. Additionally, we formed S-GEAR-2B by late-fusing two S-GEAR versions with ViT-B backbones (input $224\times224$ and $384\times384$). Despite being a late fusion (compared to RAFTformer-2B's joint architecture), S-GEAR-2B achieves a 0.5 improvement in action –- all without spatiotemporal initialization. Furthermore, S-GEAR demonstrates strong performance compared to AVT and DCR across modalities, achieving gains of 3.4 and 2.2 for action Top-5 Recall in RGB.  On the other hand, on object modality, S-GEAR shows gains of  1.7  and 4.4 (actions, nouns) and slightly trails DCR on verbs. Finally, S-GEAR remains competitive even in the flow modality.
Certainly, this detailed comparison verifies the contribution of S-GEAR in training effective anticipation models aware of action semantic interconnections, paving the way for further research.
\begin{table}[!t]
\centering
    \caption{EG results regarding Top-1 Acc. for $\tau_o=0.5$s and Top-5 Acc. for $\tau_a=1.0$s.}
    \label{tab:eg}
    \resizebox{.7\linewidth}{!}{%
    \begin{tabular}[l]{l c c c}
         \toprule
         \multirow{1}{*}{Model} & Modalities & {\begin{tabular}[c]{@{}c@{}}Top-1 Acc.\\ ($\tau_o =0.5$s)\end{tabular}} & {\begin{tabular}[c]{@{}c@{}}Top-5 Acc.\\ ($\tau_a =1$s)\end{tabular}}\\ \midrule
         RU-LSTM \cite{funari@2021rolling} & RGB+Flow & / & 66.40\\
         DCR \cite{xu@2022dcr} & RGB+Flow & / & 67.9 \\
         SRL \cite{qi@2023srl} & RGB+Flow & / & 70.7 \\
         HRO \cite{liu2022hybrid} & RGB+Flow+Obj & / & \underline{71.5} \\
         AVT \cite{girdhar@2021anticipative} & RGB & 43.0 & / \\
         AFFT \cite{zhong@2023anticipative} & RGB+Flow & 42.5 & / \\
         \textbf{S-GEAR (ours)} & RGB & \textbf{45.7} & \textbf{71.9} \\ 
         \bottomrule    
    \end{tabular}%
    }
\end{table}

\begin{table}[!t]%
\centering
    \caption{50S results on dense action anticipation. (\textit{Percentages are w.r.t. the video duration}).} 
    \label{tab:50s}
    \resizebox{.7\linewidth}{!}{%
    \begin{tabular}[r]{c | c c c c | c c c c}    
         \toprule
        $\tau_o \rightarrow$ & \multicolumn{4}{c}{20\%} & \multicolumn{4}{c}{30\%}\\ \cmidrule{1-5} \cmidrule{6-9}
        $\tau_a \rightarrow$ & 10\% & 20\% & 30\% & 50\%  & 10\% & 20\% & 30\% & 50\% \\ \midrule
        RU-LSTM \cite{funari@2021rolling} & 22.2 & 17.8 & 12.7 & 08.3 & 22.3 & 15.5 & 10.8 & 05.2 \\
        CNN model \cite{abu@2018will} & 21.2 & 19.0 & 16.0 & 09.9 & 29.1 & 20.1 & 17.5 & 10.9 \\
        Grammar-based \cite{richard2017weakly} & 24.7 & 22.3 & 19.8 & 12.7 & 29.7 & 19.2 & 15.2 & 13.1 \\
        Uncertainty \cite{abu@2019uncertainty} & 28.9 & 22.4 & 19.9 & 12.8 & 29.1 & 20.5 & 15.3 & 12.3 \\ 
        RNN \cite{abu@2018will} & 30.1 & 25.4 & 18.7 & 13.5 & 30.8 & 17.2 & 14.8 & 09.8 \\ 
        Time-Cond. \cite{ke@2019time} & 32.5 & 27.6 & 21.3 & \textbf{16.0} & 35.1 & \underline{27.1} & \textbf{22.1} & \underline{15.5} \\
        SRL \cite{qi@2023srl} & \underline{37.9} & \textbf{28.8} & \underline{21.3} & 11.1 & \underline{37.5} & 24.1 & 17.1 & 09.1 \\ 
        \textbf{S-GEAR (ours)} \cite{qi@2023srl} & \textbf{41.0} & \underline{28.5} & \textbf{21.5} & \underline{15.3} & \textbf{41.0} & \textbf{27.8} & \underline{21.4} & \textbf{16.7} \\ 
        \midrule
    \end{tabular}%
    }
\end{table}
\subsection{Comparison with the SOTA}
\noindent\textbf{Epic-Kitchens.}  Previous approaches often utilize cross-modality ensembling  \cite{dima2022rescaling,zhong@2023anticipative} or joint training \cite{funari@2021rolling,zhong@2023anticipative} for multimodal evaluation on these benchmarks. Ensembling S-GEAR across modalities, we observe significant gains. On EK55 (Table \ref{tab:ek_experiments} (c)), late-fusing our models (RGB+Obj+Flow) yields a boost of 3.5 (Top-1 Acc.) and 2.0 (Top-5 Acc.) absolute points, outperforming prior work. Similarly, on EK100 (Table \ref{tab:ek_experiments} (d)), late-fusing RGB modalities with object features leads to a 1.4 (0.8 against RAFTformer) point improvement in action Top-5 Recall. Finally, though we report EK100 test set results (Table \ref{tab:ek_experiments} (d)) and obtain competitive performances, it is crucial to note that leaderboard rankings often rely on large-scale external data or fusion across diverse models (i.e., the de-emphasized models on Table \ref{tab:ek_experiments} (d)). This makes the test set less effective for comparing the core strengths of models \cite{xu@2022dcr}. We point the reader to Appendix B.2 for details on our specific ensembling weights.

\noindent\textbf{EGTEA Gaze+.} We evaluate S-GEAR on two task on EG (Table \ref{tab:eg}). The first includes Top-1 Acc. on split-1 for $\tau_a=0.5$ where we achieve 2.5 point improvement compared to previous work. The second includes the average Top-5 Acc. across the three splits at $\tau_a=1s$ where we surprisingly improve on HRO with 0.4 points despite using only the RGB modality with our ViT-B backbone. 

\noindent\textbf{50 Salads.} Our dense anticipation experiments on the 50S (Table \ref{tab:50s}) show S-GEAR's potential for long-term and exocentric tasks. It outperforms competitors in 5/8 scenarios, with Top-1 Accuracy gains of up to 3.5, despite not being tailored for long-term anticipation like Time-Cond. \cite{ke@2019time}.

\subsection{Ablation Study} \label{ss:ablation_study}
We analyze the importance of S-GEAR's components to justify our design choices. Specifically, we investigate \textbf{(a)} the impact of architectural and training elements, \textbf{(b)} the significance of encoding semantic action relationships, \textbf{(c)} the number of prototypes for defining relative action positions, and \textbf{(d)} S-GEAR's performance for different anticipation time $\tau_a$.

\begin{table}
\centering
        \caption{Ablation study (Top-5 Recall) on EK100 validation set.}
        \resizebox{.5\linewidth}{!}{%
        \begin{tabular}{l | c c c | c c c}
             \toprule
            Settings & TCA & Sem &  PA & Verb & Noun &Action \\\midrule
            (1) Baseline & - & - & - & 30.5 & 32.6 & 15.2 \\
            (2) Sem & - & \checkmark & - & 30.7 & 35.7 & 17.8 \\
            (3) TCA & \checkmark & - & - & 31.0 & 33.9 & 16.7 \\
            (4) PA + Sem & - & \checkmark & \checkmark & \textbf{32.0} & \underline{36.2} & \underline{18.0} \\ \midrule
            (5) TCA + PA ($\rho_\ell$) & \checkmark & - & \checkmark & 30.6 & 33.3 & 17.4 \\ \midrule
            S-GEAR & \checkmark & \checkmark & \checkmark & \underline{31.1} & \textbf{37.3} & \textbf{18.3} \\ \bottomrule
        \end{tabular}%
        }
        \label{tab:ablation}
\end{table}

\begin{figure}[!h]
    \centering
     \renewcommand{\floatrowsep}{\hskip 4em}
    \begin{floatrow}
      \ffigbox[\FBwidth]{\includegraphics[width=0.45\textwidth,viewport=0 0 500 400]{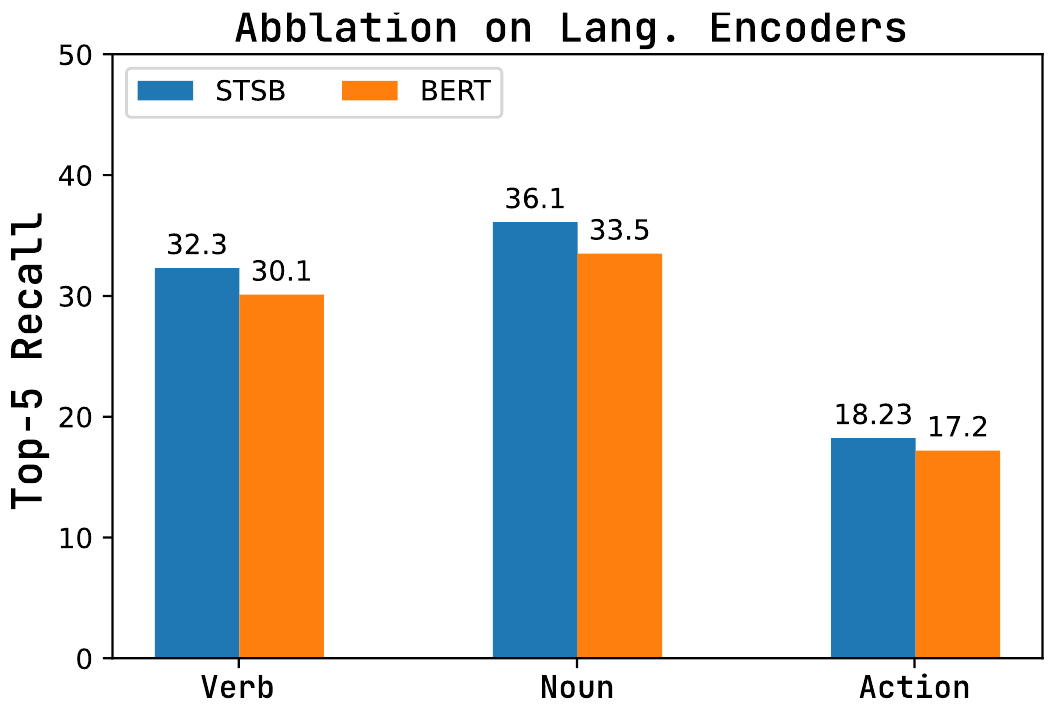}}
    {\caption{Ablation on language encoders.}\label{fig:lang_ablation}}
        \ffigbox[\FBwidth]{\includegraphics[width=0.45\textwidth,viewport=0 0 500 400]{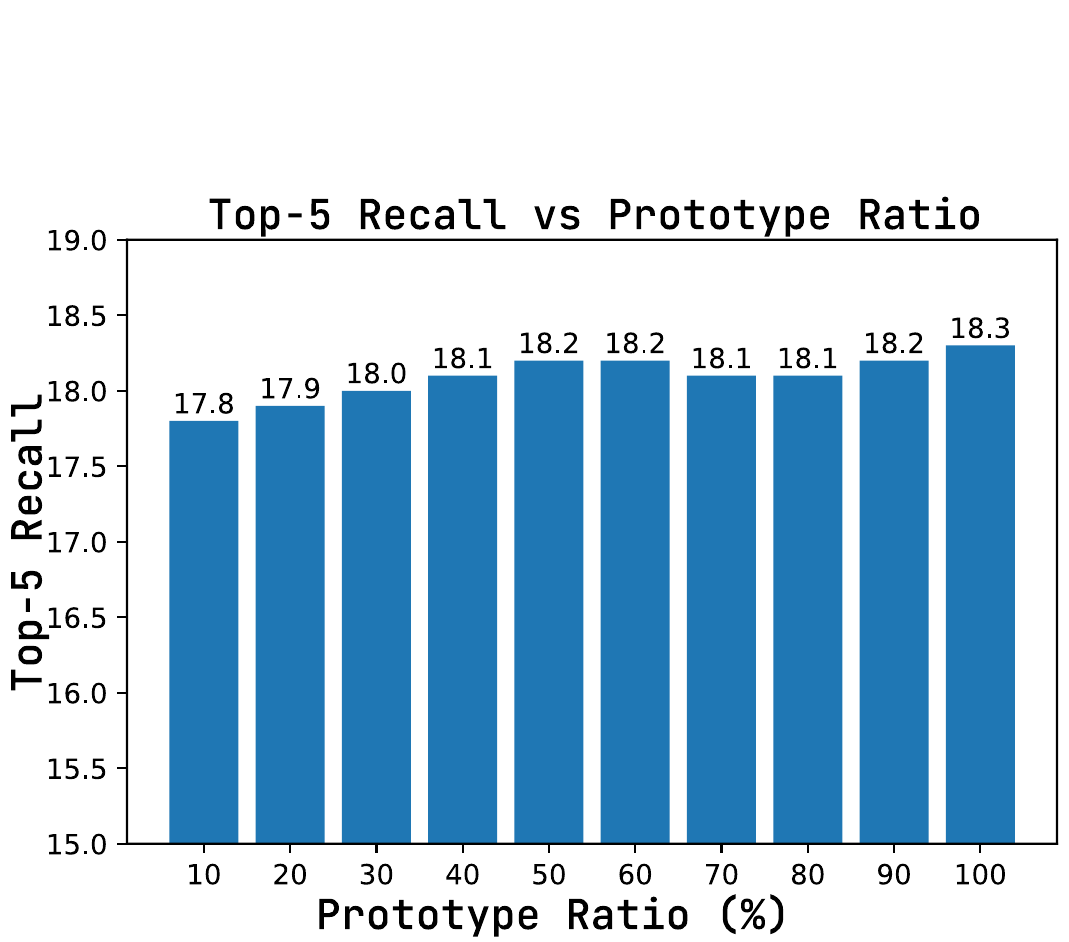}} 
        {\caption{Performance according to used prototype ratio.}\label{fig:ablation_prototype_ratio}}
    \end{floatrow}
\end{figure}

\textbf{(a)} We use EK100 (RGB) to evaluate the impact of architectural components and our prototype learning strategy (see Table \ref{tab:ablation}). We use a baseline (1) comprising a ViT-B encoder, a casual decoder, and a linear classification head similar to AVT \cite{girdhar@2021anticipative}. On top of this baseline, we switch on/off each component that comprises S-GEAR: i.e., (2) the prototype learning with semantic guidance, including the cosine attention block on the classification head, (3) the TCA block, and (4) the PA block. Note that the PA block needs prototypes; thus, in the table, we toggle the semantic column as well. While all strategies improve over (1), (2) has the most impact, adding up to 2.6 points on Top-5 Recall for action classes. Such improvements are caused by the ability of the prototype learning strategy to cluster actions that co-occur frequently. The network then uses this proximity to encode action representations aware of their exact collocation through the cosine attention block, taking hints that the next probable action can be found in its proximity in the latent space. Finally, to motivate our choice of learning visual prototypes $\rho_\upsilon$ rather than directly using language prototypes $\rho_\ell$, we rely on (5), which includes all the architecture components except the prototype learning strategy. Instead, action representations are directly aligned with fixed $\rho_\ell$. This resulted in decreased performance compared to S-GEAR.  While $\rho_\ell$ captures semantic structure, we believe it lacks the scene information crucial for accurate anticipation, such as motion and visual context. S-GEAR overcomes this limitation by learning its visual prototypes, allowing them to adapt to the specific visual cues relevant to the task.

\textbf{(b)} S-GEAR builds on the principle that semantically similar actions often co-occur, making semantic relationship encoding crucial. To ablate on the importance of such relationships, we leverage two Sentence Transformer variations from HuggingFace: ``bert-large-nli-max-token'' (BERT) and ``stsb-mpnet-base-v2'' (STSB). These models share a similar architecture but differ in training data size, with STSB being better at semantic relation extraction. Fig. \ref{fig:lang_ablation} shows that S-GEAR performs better with STSB-generated prototypes, highlighting that modeling accurate semantic interconnections gives better results.

\textbf{(c)} While prototypes are valuable, they introduce a computational cost due to their large matrix size (e.g., in EK100 with 4053 actions). In this regard, we investigate the possibility of approximating an action's relative position by comparing it to only a subset of prototypes. Experiments on EK100, using varying portions of visual prototypes (see Fig. \ref{fig:ablation_prototype_ratio}) show that we can achieve good results using only a fraction (i.e., 17.8 Top-5 Recall at 10\% vs. 18.3 Top-5 Recall at 100\%) of the prototypes while significantly reducing the number of computations (i.e., $\sim$405 instead of 4053 comparisons, entailing a 90\% drop when computing the similarity between the action representations and prototypes).

\begin{figure}[!t]
    \centering
     \renewcommand{\floatrowsep}{\hskip 1em}
    \begin{floatrow}
      \ffigbox[\FBwidth]{\includegraphics[width=0.22\textwidth,viewport=0 0 500 800]{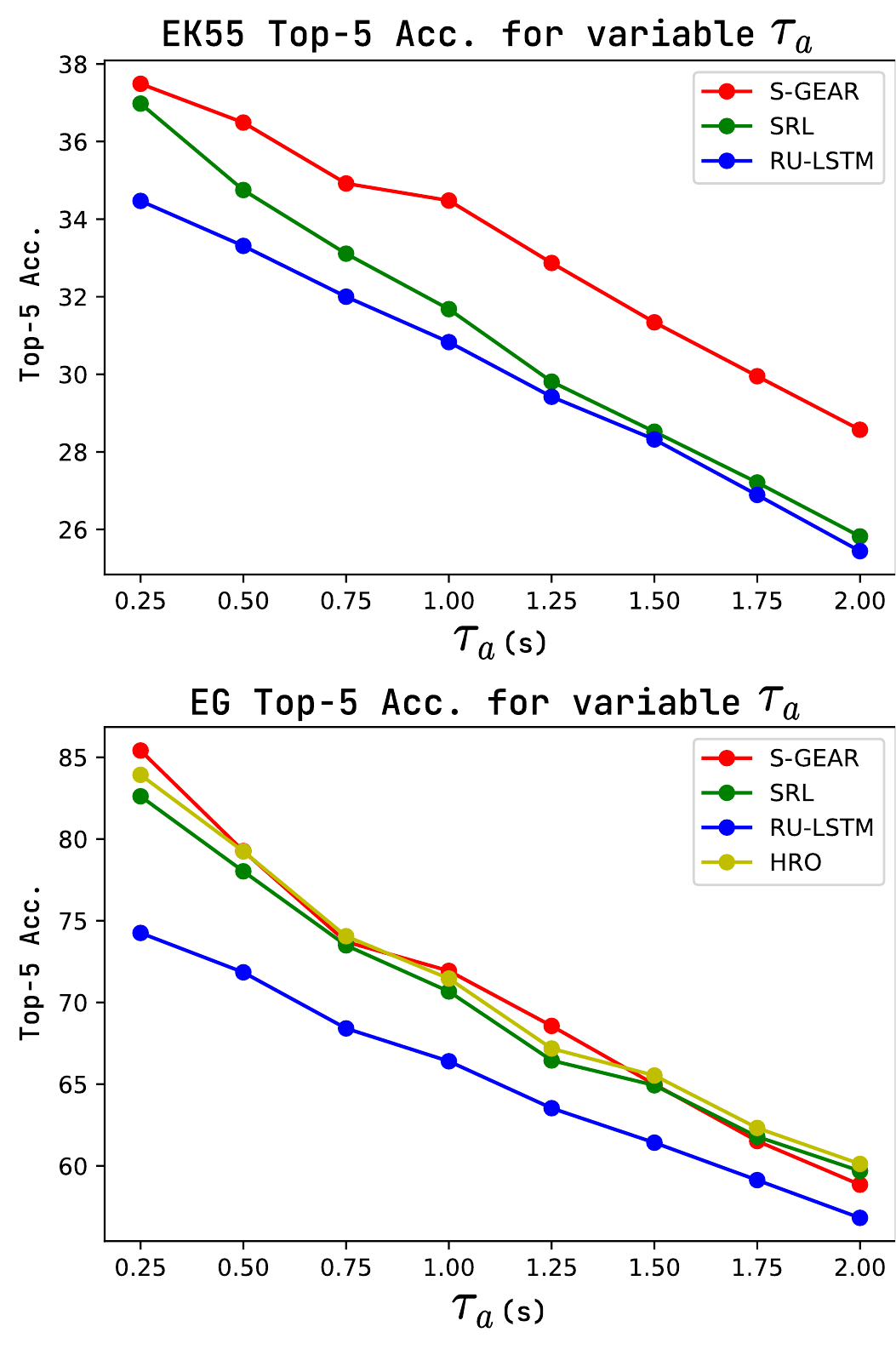}}
    {\caption{EK55 (top) and EG (bottom) Top-5 Acc. for variable $\tau_a$.}\label{fig:acc_vs_tau}}
        \ffigbox[\FBwidth]{\includegraphics[width=0.75\textwidth,viewport=0 0 3900 800]{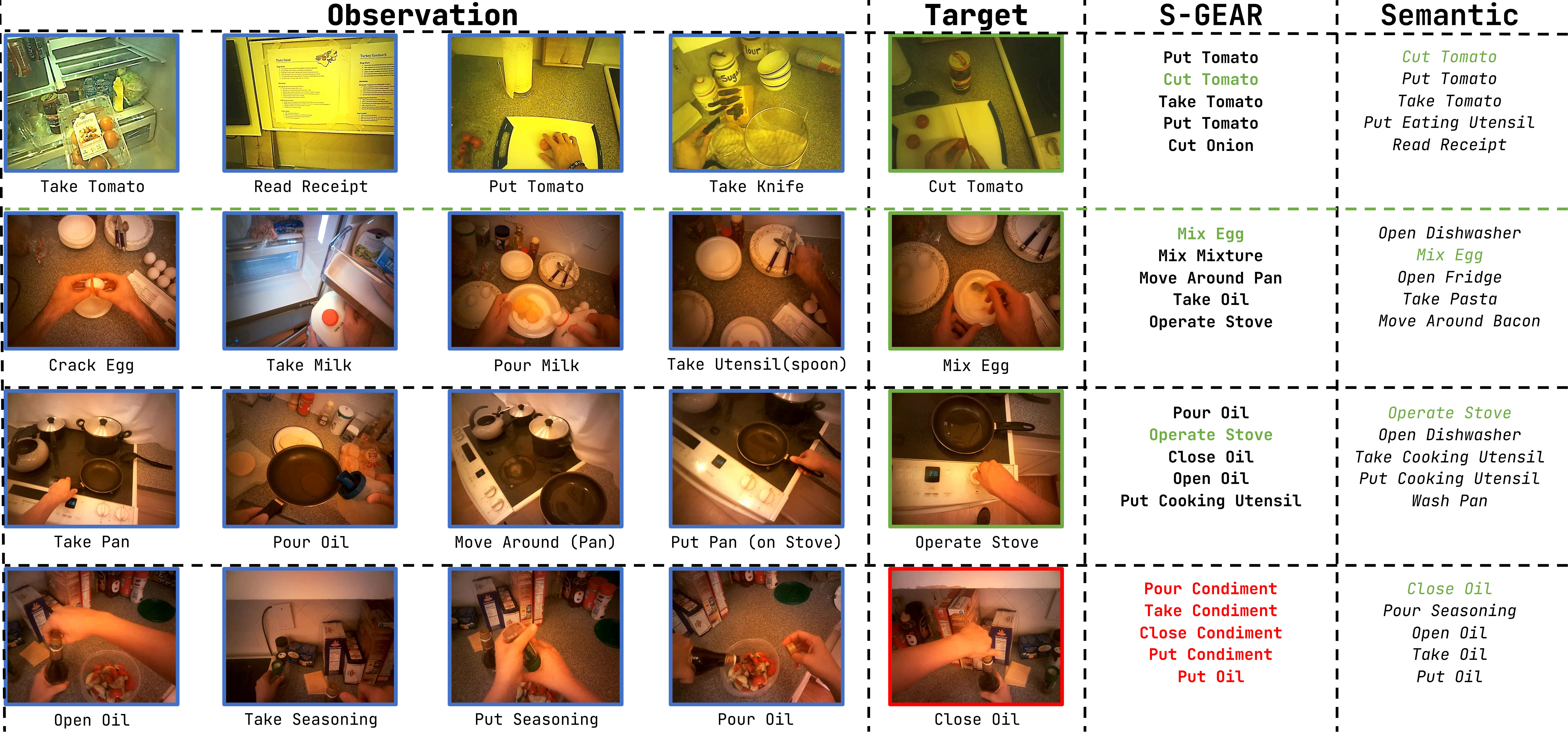}} 
        {\caption{Qualitative example of observed actions (Observation), the target activity (Target), S-GEAR's Top-5 predictions, and the Top-5 semantically similar actions with the observed sequence based on language encoding (Semantic).}\label{fig:qualitative_comparison}}
    \end{floatrow}
\end{figure}

\textbf{(d)} Finally, we evaluate the performance of S-GEAR for variable $\tau_a$. We expect the performance to drop as $\tau_a$ increases. Hence, we experiment on EK55 and EG training S-GEAR with $\tau_a=0.25$ and test its autoregressive capabilities by increasing $\tau_a$ up to 2s at inference time. We report the results in Fig. \ref{fig:acc_vs_tau}. While the performance drop is highlighted as $\tau_a\rightarrow\infty$, we notice that S-GEAR performs better than previous works on EK55. On the other hand, on EG, S-GEAR remains highly competitive, slightly trailing HRO and SRL with $\tau_a > 1$s.

\subsection{Qualitative Results}
\begin{figure}[!t]
\centering
\includegraphics[width=\textwidth,viewport=0 0 4000 960]{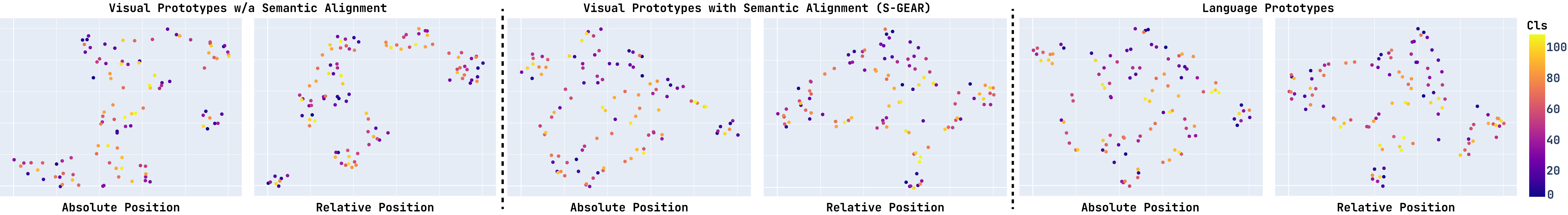}
 \caption{
Illustration (via UMAP \cite{mcinnes2018umap}) of the absolute and relative position of visual action prototypes w/a semantic alignment (left), after semantic alignment (middle), and language prototypes (right) for EG.
 }
 \label{fig:prototype_comparison}
\end{figure}

Fig. \ref{fig:qualitative_comparison} demonstrates S-GEAR's ability to anticipate future actions on the EG dataset, using $\tau_a=1$s and $\tau_o=32$s. 
Alongside S-GEAR's Top-5 predictions, we include the Top-5 semantically similar language prototypes given the observed action sequence. These examples reveal the connection between anticipation and semantics, suggesting that an alignment between the two exists. On the other hand, the last row example also highlights divergences emphasizing S-GEAR's room for semantic improvement. To further investigate the semantic alignment between S-GEAR and language prototypes, in Fig. \ref{fig:prototype_comparison}, we illustrate the geometric association learned by S-GEAR prototypes (middle) on EG, comparing it with its initial values (left) and the language prototypes (right) both in terms of absolute and relative positions. The latter is determined using cosine similarity to compare each prototype against all others. S-GEAR's prototypes demonstrate a latent space topology closer to the language prototypes than its counterpart w/o semantic alignment in terms of absolute and relative position. Such phenomenon indicates that S-GEAR can reason upon the semantic connectivity between actions, projecting contextually similar ones closer in latent space. However, S-GEAR's topology is slightly different since visual cues influence inter-prototype distances. We point the reader to Appendix B for more experimental details.

\section{Conclusion}\label{ref:conclusion}
We presented S-GEAR, a novel framework for action anticipation that leverages semantic interconnectivity between actions. S-GEAR learns visual and language prototypes that encode typical action patterns and their relationships based on contextual co-occurrences. S-GEAR transfers the geometric associations between actions from language to vision without direct alignment, creating a common communication space. S-GEAR employs a transformer-based architecture incorporating temporal context aggregation and prototype attention to enhance the action representations and predict future events. We evaluate S-GEAR on four action anticipation benchmarks, showing improved results compared to previous works. We also demonstrate that we can effectively encode semantic relationships between actions, opening new research frontiers in anticipation tasks. While S-GEAR shows promising results, its limitations include the lack of an in-built multimodal mechanism and semantic interconnections that explicitly account for occurrence order. Accounting for co-occurrence orders can reduce future prediction uncertainty, narrowing the scope of future action to those likely to follow the observed sequence.  We will address these limitations in future work.

\section*{Acknowledgments}

We would like to express our sincere gratitude to Professor Giovanni Maria Farinella and Professor Antonino Furnari from the University of Catania for their invaluable guidance, support, and insightful feedback. Their expertise and encouragement were instrumental in the successful completion of this work.

\bibliographystyle{unsrt}  
\bibliography{main}


\begin{appendix}

\section{S-GEAR's Details}\label{sss:architecture_details}
The main paper gives a general overview of S-GEAR architecture, which comprises a visual encoder, the temporal content aggregation module, the prototype attention module, and the causal transformer decoder. While the visual encoder \cite{dosovitskiy2021ViT,ashish@2017attention} -- a standard ViT -- and the temporal decoder \cite{ashish@2017attention,xu@2022dcr,girdhar@2021anticipative}-- causal transformer based on masked-attention -- are well-known architectures in the literature, we specifically tailor the two intermediate modules to serve our purposes. We provide the details of such blocks in Sec. \ref{sss:contextual_temporal_encoding} and \ref{sss:cross_modality_fusion}. In Sec. \ref{sss:prot_init}, give details regarding the visual prototype initialization. Finally, in Sec. \ref{ss:feat_train}, we give details regarding the training strategy used to train S-GEAR with pre-extracted features.
\begin{figure}[!h]
    \centering
    \includegraphics[width=\textwidth,viewport=0 0 300 125]{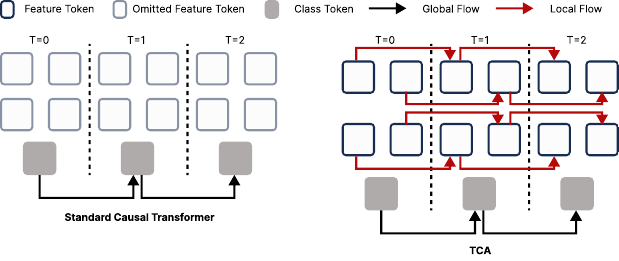}
    \caption{Standard causal transformer flow vs. TCA flow.}
    \label{fig:tca_flow}
\end{figure}

\subsection{Temporal Context Aggregation}\label{sss:contextual_temporal_encoding}

Sequential models like LSTMs and causal transformers excel at handling temporal frame sequences. However, relying on class tokens, they prioritize global information \cite{xu@2022dcr,girdhar@2021anticipative} and neglect spatial cues. To illustrate the difference between the standard causal transformer and our proposed \textit{Temporal Context Aggregator} (TCA), we provide the reader with Fig. \ref{fig:tca_flow}. Here, the left-hand side shows the workflow of a standard causal transformer applied on a sequence of ViT frame features composed of \textit{local} feature tokens and a \textit{global} class token. In this scenario, the causal transformer omits the local information and only propagates the global information in time to create causal representations \cite{girdhar@2021anticipative,xu@2022dcr}. Because local tokens encode specific scene details within different regions, not propagating their information hinders the model from understanding scene dynamics (e.g., how an object's location changes as a particular action progresses). Therefore, we design the TCA $\varphi$ block. TCA extends the information flow by propagating global and local tokens across time (see Fig. \ref{fig:tca_flow} right), building causal representations considering scene dynamics at a finer spatial scale.

\begin{figure}[!t]
\centering
\includegraphics[width=\textwidth,viewport=0 0 2250 1700]{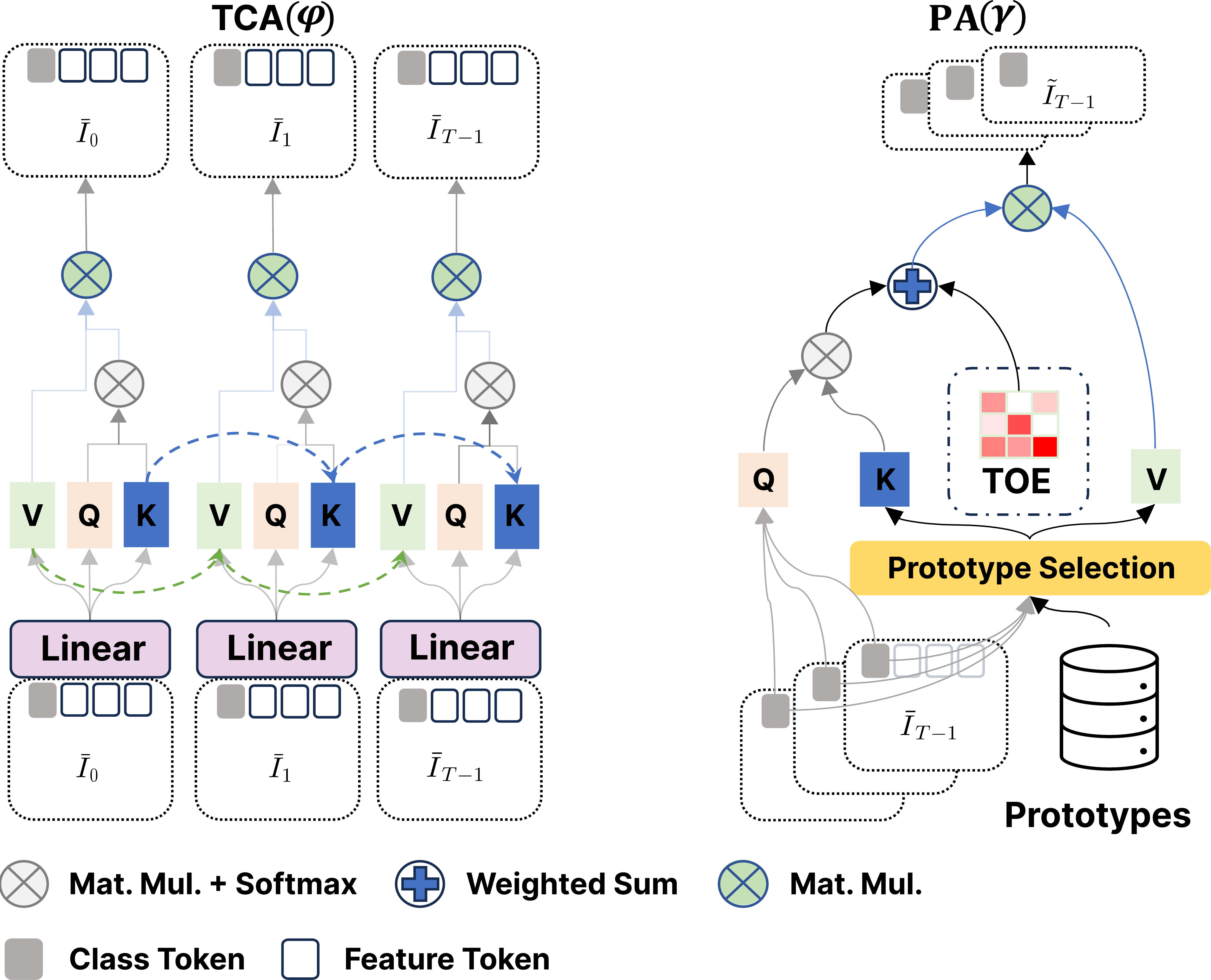}
 \caption{
The TCA block (left) generates detailed and sequential representations by aggregating past observation values $V$ and keys $K$ into current ones. It follows a standard attention mechanism to capture temporal dependencies. The PA block (right) performs cross-modality attention to aggregate selected prototypes from visual queries and uses a TOE weight matrix to encode temporal awareness. Note that the red gradient in TOE represents the magnitude of learnable weights.
}
 \label{fig:tca_pa}
\end{figure}

TCA builds on the attention mechanism and processes intermediate features $I_t$. Thus, $I_t$ undergoes linear processing to generate the query ($Q_t$), key ($K_t$), and value ($V_t$) vector representations. Afterward, as shown in Fig. \ref{fig:tca_pa} (left), the TCA uniquely aggregates keys and values from past frames to subsequent ones before computing the attention matrix. This approach enables the queries of each frame to access a rich set of keys and values infused with comprehensive spatiotemporal information about past contexts, enabling better temporal dependency \cite{wu@2022memvit}. Specifically, the $K_t$ and $V_t$ vectors are augmented as in Eqns. \ref{eq:q_agreggation} and \ref{eq:k_aggregation}, respectively:
\begin{equation}\label{eq:q_agreggation}
    \hat{K}_{t} =
    \begin{cases}
     K_t & \text{if}\  t=0 \\
     \delta(K_t, \alpha_{t-1}\cdot \hat{K}_{t-1}) & \text{otherwise}
    \end{cases},
\end{equation}
\begin{equation}\label{eq:k_aggregation}
    \hat{V}_{t} =
    \begin{cases}
     V_t & \text{if}\  t=0 \\
     \delta(V_t, \alpha_{t-1}\cdot \hat{V}_{t-1}) & \text{otherwise}
    \end{cases},
\end{equation}
where $\hat{K}_{t}$ and $\hat{V}_{t}$  are the augmented keys and queries of frame $f_t$,  $\alpha_{t-1}$ is a learnable weight parameter that balances the quantity of the information transmitted from past observations, and $\delta$ is a permutation invariant aggregation function. Note that we tried different functions for $\delta$ (e.g., cumulative-max), but through empirical analyses, we chose summation.

Once these vectors are obtained, the computation proceeds with the standard self-attention operations. Precisely, $Q_t$ and $\hat{K}_t$ compute the attention scores through scaled matrix multiplication, then normalized into $[0,1]$ weights through a softmax function. Afterwards, the weights aggregate information between the augmented feature tokens of $\hat{V}_t$ and create causal representations $\overline{I}_t$. 
Formally, the procedure can be defined as follows:
\begin{equation}\label{eq:TCA_attn}
    \overline{I}_t  = \text{softmax}\bigg(\frac{Q_t \hat{K}_t^\top}{\sqrt{d}}\bigg) \hat{V}_t.
\end{equation}
Now, the class token representations $\overline{I}_t^0$ of $\overline{I}_t$ encodes the global representation of frame $t$ enhanced by detailed contextual information of the past frames.

\subsection{Prototype Attention}\label{sss:cross_modality_fusion}
Recall that S-GEAR encodes semantic relationships between actions. To help the encoding of such relationships, we integrate a \textit{Prototype Attention} (PA) block $\gamma$ detailed in Fig. \ref{fig:tca_pa} (right). The PA module helps the network learn meaningful representations by incorporating semantic information from the visual prototypes. PA has two stages: \textbf{(1)} selecting the prototypes and \textbf{(2)} modeling the relationship between features and prototypes.

Similar to TCA, we build PA upon the attention mechanism, giving in input both the class tokens from the intermediate encodings generated by ViT -- i.e., $I^0=\{I_0^0, I_1^0,\dots, I_{T-1}^0\}$ and the visual prototypes $\rho_\upsilon$. We rely on the relative similarities between actions to address \textbf{(1)}. Specifically, we begin by calculating the cosine similarity between each $I_t^0$ and the visual prototypes, obtaining the relative representation vector $\mathbf{r}^{I^0_t}$ of frame $t$ as in Eq. \ref{eq:sim_vis_1}:
\begin{equation}\label{eq:sim_vis_1}
   \mathbf{r}^{I^0_t} = cos(I_t^0, \rho_\upsilon).
\end{equation}
Then, we select the top $k$ most similar prototypes for each feature vector for the remaining calculations. However, for simplicity, let us assume we select the most similar prototype for each feature vector. After acquiring the estimated prototypes, PA addresses \textbf{(2)} by modeling their relationship with the feature encodings using the attention mechanism. In this case, the set of prototypes represents both the key ($K$) and value ($V$) vectors. Conversely, the query ($Q$) vector is derived from $I^0$. The first step of the relationship modeling is the computation of the attention scores $W_a$
through a scaled matrix multiplication between $Q$ and $K$ as in Eq. \ref{eq:attn_score}:
\begin{equation}\label{eq:attn_score}
    W_a = \frac{QK^\top}{d}.
\end{equation}
Following the standard attention procedure, the next step in the attention process should be normalizing and applying $W_a$ to $V$ and having the output features. However, the selected prototypes do not have temporal continuity like the sequential frames and contradict the temporal causality built from TCA when the fusion occurs (see Sec. 3.1 in MP). Inspired by \cite{huang2022encoding}, we introduce a \textit{Temporal Order Encoding} (TOE) weight vector shaped as a Toeplitz matrix to model the temporal order between elements of $V$. We provide the reader with an example to illustrate Toeplitz matrices and their unique structure. Here, we show a 5-element TOE as a $3\times 3$ Toeplitz matrix $\Delta$ as in Eq. \ref{eq:toeplitz}:
\begin{equation}\label{eq:toeplitz}
\Delta = 
\left(
    \begin{array}{ccc}
         w_{0}& w_{1} &w_{2}  \\
         w_{3}& w_{0} &w_{1}  \\
         w_{4}& w_{3} &w_{0}  \\ 
    \end{array}
\right),
\end{equation}
where $w_i$ represents the $i^{\text{th}}$ weight from the TOE for $i \in \{0, 1, ..., 4\}$. Notice that a single weight represents each diagonal. Additionally, with a $3 \times 3$ matrix, we can model the temporal relationships of a sequence of three elements. Hence, for a sequence of $T$ elements like $V$, we need a $T\times T$ Toeplitz matrix built from a ($2T-1$)-element TOE. Generalizing, the $T\times T$ $\Delta$ functionality allows PA to model the relative temporal position or order between elements of $V$ when aggregating features. To apply $\Delta$ to $V$, we first sum $\Delta$ with the normalized $W_a$ and then perform a matrix multiplication between the resulting matrix and $V$ as in Eq. \ref{eq:v_times_wa}:
\begin{equation}\label{eq:v_times_wa}
    \Tilde{I} = (\beta(\text{softmax}(W_a)) + (1-\beta)\Delta)V,
\end{equation}
where $\beta$ is a learnable scaling factor that balances the sum between $W_a$ and $\Delta$. $\Tilde{I} \in \mathbb{R}^{T\times d}$ now represents the selected visual prototypes fused with the current context and with encoded relative temporal awareness. 


\subsection{Prototype Initialization} \label{sss:prot_init}

We initialize our visual prototypes $\rho_\upsilon$ using action samples generated by the proposed architecture (detailed in Sec. 3.1 in MP). First, we train the network (PA omitted) for action recognition on EK100 for the egocentric tasks and 50S for the exocentric tasks. Then, without fine-tuning, we extract multiple representations for each action class present in a dataset. The average of these representations becomes the ``typical action sample'' used to initialize the corresponding class prototype.  We also experimented with other initialization methods like random values and using the language prototypes $\rho_\ell$ as initial values. However, such methods did not provide significant improvements. Thus, the results we report in MP rely on the first initialization approach.

\begin{figure}[!ht]
    \centering
    \includegraphics[width=\textwidth,viewport=0 0 600 370]{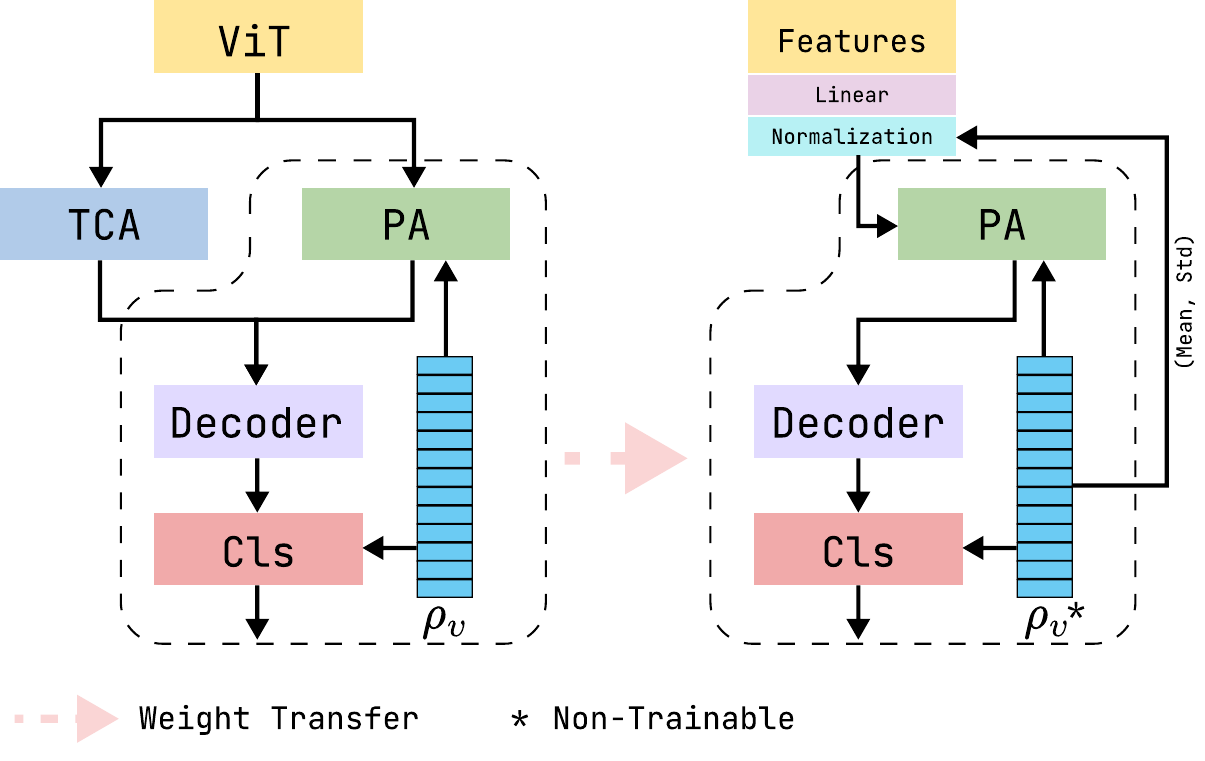}
    \caption{Training S-GEAR with pre-extracted features.}
    \label{fig:feature_training}
\end{figure}

\subsection{Training S-GEAR With Pre-extracted Features} \label{ss:feat_train}

S-GEAR's modular design allows it to work with different visual backbones, although it's primarily designed for end-to-end training with a ViT architecture \cite{dosovitskiy2021ViT}. To ensure a fair comparison with the SOTA \cite{funari@2021rolling,xu@2022dcr,girdhar@2021anticipative}, we also train our network using pre-extracted features from TSN, Faster R-CNN (FRCNN) and irCSN backbones provided by Furnari et al. \cite{funari@2021rolling} and Girdhar et al. \cite{girdhar@2021anticipative}, respectively.

Our training process (see Fig. \ref{fig:feature_training}) is built upon aligning the pre-extracted feature distribution with $\rho_\upsilon$ learned from S-GEAR in its end-to-end training with ViT. These prototypes already capture the desired structure of the latent space. By aligning with them, we simplify training when we lack the variability introduced by typical video preprocessing techniques (i.e., random cropping or flipping). To this end, given pre-extracted visual features $\chi_t$  $\forall t \in [0, T-1]$, we first apply a linear transformation and then normalize them using the mean $\mu_{\rho_\upsilon}$ and standard deviation $\sigma_{\rho_\upsilon}$ from the learned prototypes as in Eq. \ref{eq:normalization}:
\begin{equation}
    I_t = \frac{\text{lin}(\chi_t) - \mu_{\rho_\upsilon}}{\sigma_{\rho_\upsilon}}.
    \label{eq:normalization}
\end{equation}
Hence, leveraging the pre-trained weights of PA, causal decoder, and classification head from the end-to-end training, we fine-tune S-GEAR to adapt to the new visual features $I_t$. During this process, $\rho_\upsilon$ remains unchanged.

\section{Experiments Extension}\label{sss:extended_experiments}
Here, we provide additional details regarding our experiments. In Sec. \ref{ss:loss_w}, we provide details regarding the weights used for each loss function to train S-GEAR for each dataset. In Sec. \ref{ss:ablation_tau}, we provide the results supporting the ablation study on the anticipation time $\tau_a$ from Sec. 4.6 in the MP. In Sec. \ref{ss:ensamble}, we give details regarding the ensemble setup of our models used to obtain the multimodal results on EK55/100 from the MP. In Sec. \ref{ss:mm_ablation}, we try different combinations of backbones and modalities on EK100's validation set. Finally, in Sec. \ref{ss:semanticity}, we extend Sec. 4.7 from the MP and compare the semanticity learned from $\rho_\upsilon$ and that encoded from $\rho_\ell$.

\subsection{Composed Loss Weights} \label{ss:loss_w}

Here, we provide the specific weights for each loss term introduced in Sec. 3.3 of the MP. Table \ref{tab:weights_loss} details these weights. Notice that we combine $\Lagr_{Sem}$ and $\Lagr_{Reg}$ due to their similar purpose in MP. However, they are weighted individually during optimization to account for their different magnitudes. Importantly, we use the same weights for both EK100 and EK55 datasets, regardless of whether training end-to-end with ViT features or using pre-extracted features. The sole exception is  $\Lagr_{Sem}$ because visual prototypes remain frozen in the pre-extracted feature training scenario (see Sec. \ref{ss:feat_train}).

\begin{table}[!t]
    \centering
    \caption{Weights associated with individual loss terms to train S-GEAR for each task.}
    \begin{tabular}{c c c c c c c c}
        \toprule
        Dataset & & $\Lagr_{Sem}$ & $\Lagr_{Reg}$ & $\Lagr_{Cls}$ & $\Lagr_{Past}$ & $\Lagr_{Feat}$ \\ \midrule
        EK100 && 4.0 & 1.0 & 1.0 & 1.0 & 1.0 \\
        EK55 && 2.0 & 1.0 & 1.0 & 1.0 & 1.0 \\
        EG && 2.0 & 1.0 & 1.0 & 0.1 & 1.0 \\
        50S && 1.0 & 0.1 & 1.0 & 0.1 & 1.0 \\
        \bottomrule
    \end{tabular}
    \label{tab:weights_loss}
\end{table}

\subsection{Ablation on anticipation time}\label{ss:ablation_tau}
Here, we report the results of experiments in Sec. 4.6 of MP. Specifically, the results correspond to Fig. 5 of MP exploring Top-5 Acc. on EK55 and EG for $\tau_a$ spanning from 0.25s to 2.0s.
\begin{table}[!t]
    \centering
    \caption{Results on the EK55 validation set regarding Top-5 Accuracy at different Anticipation Time-Steps averaged across the three official splits. All results in this table are obtained using only RGB modality.}
    \begin{tabular}{l c c c c c c c c}
         \toprule
         \multirow{2}{*}{Model} & \multicolumn{8}{c}{Top-5 Accuracy \% at different $\tau_a$ (s)} \\ \cmidrule{2-9}
         &  $2.0$ & $1.75$ & $1.5$ & $1.25$ & $1.0$ & $0.75$ & $0.5$ & $0.25$\\ \midrule
         FN \cite{geest@2018modeling} & 23.47 & 24.07 & 24.68 & 25.66 & 26.27 & 26.87 & 27.88 & 28.96 \\
         RL \cite{ma@2016learning} & \underline{25.95} & 26.49 & 27.15 & 28.48 & 29.61 & 30.81 & 31.86 & 32.84 \\
         RU-LSTM \cite{funari@2021rolling} &  25.44 & 26.89 & 28.32 & 29.42 & 30.83 & 32.00 & 33.31 & 34.47 \\
         SRL \cite{qi@2023srl}&  25.82 & \underline{27.21} & \underline{28.52} & \underline{29.81} & 31.68 & \underline{33.11} & \underline{34.75} & \underline{36.89} \\
         \textbf{S-GEAR (ours)} &  \textbf{28.57} & \textbf{29.95} &\textbf{31.34} & \textbf{32.87} & \textbf{34.48} & \textbf{34.92} & \textbf{36.49} & \textbf{37.49} \\ 
         \bottomrule    
    \end{tabular}%
    \label{tab:1}
\end{table}

\begin{table}[!t]
    \centering
    \caption{Results on the EGTEA Gaze+ validation set regarding Top-5 Accuracy at different Anticipation Time-Steps averaged across the three official splits.}
    \begin{tabular}{l c c c c c c c c}
         \toprule
         \multirow{2}{*}{Model} & \multicolumn{8}{c}{Top-5 Accuracy \% at different $\tau_a$ (s)} \\ \cmidrule{2-9}
         & $2.0$ & $1.75$ & $1.5$ & $1.25$ & $1.0$ & $0.75$ & $0.5$ & $0.25$\\ \midrule
         FN \cite{geest@2018modeling} & 54.06 & 54.94 & 56.75 & 58.34 & 60.12 & 62.03 & 63.96 & 66.45 \\
         RL \cite{ma@2016learning} & 55.18 & 56.31 & 58.22 & 60.35 & 62.56 & 64.65 & 67.35 & 70.42 \\
         RU-LSTM \cite{funari@2021rolling} & 56.82 & 59.13 & 61.42 & 63.53 & 66.40 & 68.41 & 71.84 & 74.25 \\
         SRL \cite{qi@2023srl} & \underline{59.69} & \underline{61.79} & 64.93 & 66.45 & 70.67 & 73.49 & 78.02 & 82.61 \\
         HRO \cite{liu2022hybrid} & \textbf{60.12} & \textbf{62.32} & \textbf{65.53} & \underline{67.18} & \underline{71.46} & \textbf{74.05} & \underline{79.24} & \underline{83.92} \\
         \textbf{S-GEAR (ours)} & 58.85 & 61.52 &\underline{64.99} & \textbf{70.56} & \textbf{71.93} & \underline{73.71} & \textbf{79.26} & \textbf{85.41} \\ \bottomrule    
    \end{tabular}
    \label{tab:4}
\end{table}

\subsection{Multimodal Ensemble} \label{ss:ensamble}

For EK100's validation set, we evaluate three S-GEAR versions with varying backbone combinations. Firstly, S-GEAR uses late fusion\footnote{Weighted Combination of predictions from different models and modalities.} of ViT-based RGB S-GEAR (weight: 2.5) and FRCNN object features (weight: 0.5).  S-GEAR-2B adds ViT$\downarrow$-based RGB S-GEAR (weight: 1.5) to the previous fusion. Finally, S-GEAR-4B combines all RGB S-GEAR variants (ViT, ViT↓, TSN, irCSN with weights 2.5:1.5:1:1) and FRCNN object features (weight: 0.5). The same weight combinations apply for the EK100 test set. On the EK55 validation set, we late fuse  ViT, irCSN, and TSN-based RGB S-GEAR (all weighted 1.5), along with TSN flow features (weight: 1) and FRCNN object features (weight: 1).  Note that except for the ViT backbones, all other RGB and modality features are pre-extracted as provided in \cite{funari@2021rolling,girdhar@2021anticipative}.

\subsection{Fine-grained Multimodal Ablation} \label{ss:mm_ablation}
\begin{table}[!t]
    \centering
    \caption{Ablation of different RGB backbones and modalities on EK100 validation set. ViT$\downarrow$ represents ViT with an input size of 224$\times$224. All backbone and modalities are combined through late fusion.}
    \begin{tabular}{c c c c c c c}
        \toprule
        RGB Backbones & & Modalities & & Verb & Noun & Action \\ \midrule
         ViT & & RGB + Obj & & 29.5 & 37.8 & 18.9 \\
         ViT & & RGB + Flow & & 30.2 & 35.4 & 18.4 \\
         ViT & & RGB + Obj + Flow & & 29.6 & 37.4 & 18.9 \\ \midrule
         ViT + ViT$\downarrow$ & & RGB + Obj & & 30.5 & \textbf{38.4} & \underline{19.6} \\ 
         ViT + ViT$\downarrow$ & & RGB + Flow & & \textbf{30.8} & 36.2 & 19.5 \\ 
         ViT + ViT$\downarrow$ & & RGB + Obj + Flow & & 30.2 & 36.9 & 19.6 \\ \midrule
         ViT + TSN & & RGB + Obj & & 30.4 & 38.1 & 19.5 \\          
         ViT + TSN & & RGB + Flow & & \underline{30.8} & 36.6 & 19.2 \\ 
         ViT + TSN & & RGB + Obj + Flow & & 29.7 & 37.8 & 19.4 \\ \midrule
         ViT + irCSN & & RGB + Obj & & 30.5 & \underline{38.2} & 19.6 \\          
         ViT + irCSN & & RGB + Flow & & 29.5 & 37.0 & 19.1 \\ 
         ViT + irCSN & & RGB + Obj + Flow & & 29.4 & 37.6 & 19.6 \\ \midrule
         ViT + ViT$\downarrow$ + irCSN + TSN & & RGB + Obj & & 30.2 & 37.0 & \textbf{19.9} \\          
         ViT + ViT$\downarrow$ + irCSN + TSN & & RGB + Flow & & 30.4 & 36.6 & 19.5 \\ 
         ViT + ViT$\downarrow$ + irCSN + TSN & & RGB + Obj + Flow & & 29.9 & 37.3 & 19.6 \\ \midrule
    \end{tabular}
    \label{tab:mm_ensamble}
\end{table}

We present results from combining different backbones and modalities on the EK100 validation set (see Table \ref{tab:mm_ensamble}). We use late fusion with the following backbone weights: ViT (weight: 2.5), ViT↓ (weight: 1.5), TSN (weight: 1.0), irCSN (weight: 1.0), and equal weights (weight: 0.5) for Object and Flow features.  We report results for various combinations, ranging from models using a single backbone to those combining up to four. We can notice that combining RGB and object modalities provides the best overall results regarding Top-5 Recall, which motivates our choice to exclude flow from the final ensemble model.

\subsection{Closer Look at Semanticity} \label{ss:semanticity}

\begin{figure}[!t]
    \centering
    \includegraphics[width=\textwidth,viewport=0 0 2650 1100]{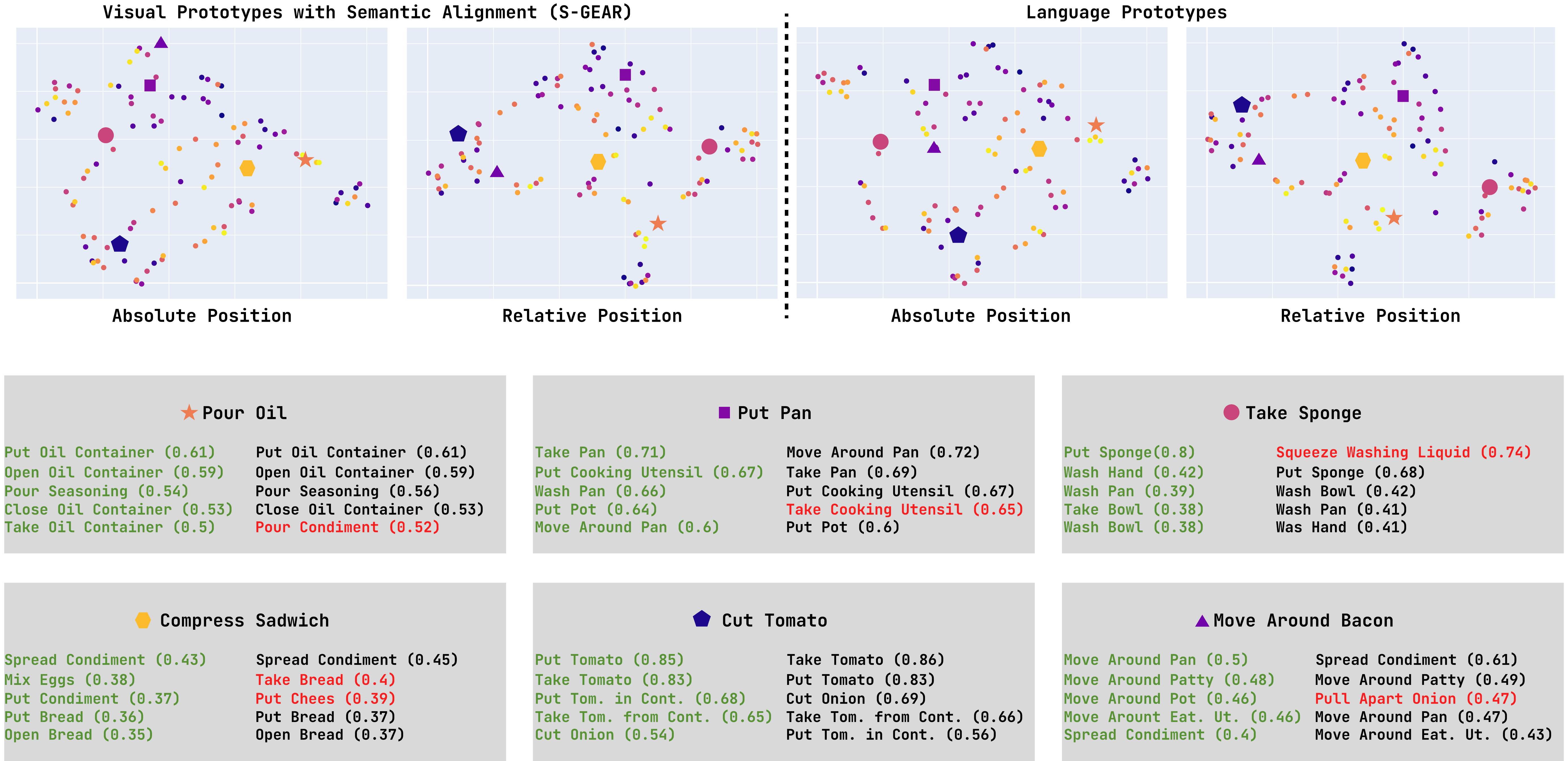}
    \caption{(best viewed in color) Fine-grained semanticity comparison.}
    \label{fig:supp_semanticiy_comparison}
\end{figure}

In Sec. 4.7 of MP, we graphically show that the latent space topology defined from S-GEAR's learned prototypes is similar to the one defined from language prototypes but not perfectly aligned due to the influence of visual cues. Such resemblance in latent space geometry suggests that S-GEAR can semantically reason regarding action associations and consider scene information simultaneously. To dive deeper into this aspect, here we consider a group of randomly selected reference actions -- i.e., \textit{Pour Oil}, \textit{Put Pan}, \textit{Take Sponge}, \textit{Compress Sandwich}, \textit{Cut Tomato}, and \textit{Move Around Bacon} -- and analyze the prototypes found in their proximity. Graphically, we show these comparisons in Fig. \ref{fig:supp_semanticiy_comparison}. The grey boxes report the top-5 most similar actions for each reference action (top bold string). Green actions (text on the left in each grey box) represent the most similar actions in the language space. Black actions (right side of the grey boxes) highlight alignment between vision and language, whereas red actions state that there is a mismatch between the two. Each action contains the cosine similarity between it and the reference one (see values inside the brackets). Notice that actions are associated with the same activities in both spaces regarding action classes and the magnitude of their cosine similarity. Nevertheless, in 5/6 reported cases, we have at least one divergence between the actions and the reference one. This phenomenon re-emphasizes the divergence of visual prototypes influenced by visual cues and the co-occurrences of actions in videos. Interestingly, even divergent actions mostly have reasonable connections (i.e., they are likely to co-occur) with the reference action. This underscores S-GEAR's prototypes' ability to keep their semantic composure and account for the co-occurrence of actions influenced by the observed action segments\footnote{Notice that the observed segments come from the EG dataset}. Notice that in the case of \textit{Take Sponge} as a reference action, the divergence between S-GEAR and language prototypes is the action \textit{Squeeze Washing Liquid}, which is probably the most likely action to co-occur with \textit{Take Sponge} in a kitchen scenario. Additionally, it shows that with the proposed method, a perfect alignment between two spaces cannot be obtained as long as the task is bounded to a given dataset. However, in such cases, as we showed in Sec. 4.6  of MP (5th setup from Table 3), relying completely on global semantics (i.e., language prototypes) is less beneficial for action anticipation than merging it with the visual cues (S-GEAR) due to the importance of motion and scene composition in suggesting possible future actions.

\end{appendix}

\end{document}